%% file: main.tex
\documentclass[runningheads]{llncs}

\usepackage{eccv}

\usepackage{algorithm,algpseudocode}

\usepackage{eccvabbrv}

\usepackage[pdftex]{graphicx}

\usepackage{booktabs}
\usepackage{wrapfig}
\usepackage{pifont}
\usepackage{mathtools}
\usepackage{xcolor,colortbl}
\newcommand{\xmark}{\ding{55}}
\usepackage[accsupp]{axessibility}  % Improves PDF readability for those with disabilities.

\usepackage{hyperref}

\usepackage{orcidlink}

\begin{document}

\title{Diffusion Prior-Based Amortized Variational Inference for Noisy Inverse Problems} 

\titlerunning{Diffusion prior-based Amortized Variational Inference (DAVI)}

\author{Sojin Lee\inst{1}\thanks{equal contribution,  $^\dagger$corresponding author}\orcidlink{0000-0001-9198-8437} \and
Dogyun Park\inst{1}$^{\star}$\orcidlink{0009-0009-1156-7559} \and
Inho Kong\inst{1}\orcidlink{0009-0003-7567-6558} \and
Hyunwoo J. Kim\inst{1}$^\dagger$\orcidlink{0000-0002-2181-9264}}

\authorrunning{S.~Lee et al.}

\institute{
Korea University, Seoul, Republic of Korea \\
\email{\{sojin\_lee,gg933,inh212,hyunwoojkim\}@korea.ac.kr}
}

\input{Sections/sojin_def}
\maketitle

\input{Sections/0_abstract}
\input{Sections/1_introduction}
\input{Sections/2_relatedworks}
\input{Sections/3_preliminary}
\input{Sections/4_method}
\input{Sections/5_experiments}
\input{Sections/6_conclusion}

\par\vfill\par
% Now we have reached the maximum length of an ECCV \ECCVyear{} submission (excluding references).
% References should start immediately after the main text, but can continue past p.\ 14 if needed.
% \clearpage  % TODO REVIEW/FINAL: This \clearpage needs to be removed from both review and camera-ready versions.
\section*{Acknowledgement}
This work was supported by ICT Creative Consilience Program through the Institute of Information \& Communications Technology Planning \& Evaluation (IITP) grant funded by the Korea government (MSIT)(IITP-2024-2020-0-01819, 10\%), the National Research Foundation of Korea (NRF) grant funded by the Korea government (MSIT) (NRF-2023R1A2C2005373, 30\%), the National Supercomputing Center with supercomputing resources including technical support (KSC-2023-CRE-0325, 30\%), and Artificial intelligence industrial convergence cluster development project funded by the Ministry of Science and ICT (MSIT, Korea)\&Gwangju Metropolitan City (30\%).

\bibliographystyle{splncs04}
\bibliography{main}

\clearpage
\appendix
\begin{center}
	\textbf{\Large Supplementary Material}
\end{center}
\input{Section_supple/A_derivations}
\input{Section_supple/B_additional_anlaysis}
\input{Section_supple/C_experimental_details}
\input{Section_supple/D_more_relatedworks}
\input{Section_supple/E_qualitative_results}

\end{document}

%% file: Sections/sojin_def.tex
\newcommand{\mathcolorbox}[2]{\colorbox{#1}{$\displaystyle #2$}}

\def\wbst{\mathbf{w}^{\ast}}
\newcommand{\Vc}{\mathcal{V}}

\newcommand{\sjcamera}[1]{\textcolor[rgb]{0.8,0.1,0.7}{#1}}

\newcommand{\sj}[1]{\textcolor[rgb]{0,0,0}{#1}}
\newcommand{\hjk}[1]{\textcolor[rgb]{0,0,0}{#1}}

\newcommand{\invD}[1]{\left (D^{(#1)} \right )^{-1}}
\newcommand{\Ll}{\mathcal{L}}
\newcommand{\Uu}{\mathcal{U}}
\newcommand{\Ss}{\mathcal{S}}
\newcommand{\Ee}{\mathcal{E}}
\newcommand{\Vv}{\mathcal{V}}
\newcommand{\Gg}{\mathcal{G}}
\newcommand{\Ff}{\mathcal{F}}
\newcommand{\kcompG}{\mathbf{G}^{(K)}}
\newcommand{\nkcompG}{\mathbf{G}^{(NK)}}
\newcommand{\lcompG}{\mathbf{G}^{(L)}}
\newcommand{\onecompG}{\mathbf{G}^{(1)}}
\newcommand{\Cc}{\mathcal{C}}
\newcommand{\Mc}{\mathcal{M}}
\newcommand{\Lc}{\mathcal{L}}
\newcommand{\Tc}{\mathcal{T}}
\newcommand{\Oc}{\mathcal{O}}
\newcommand{\Rc}{\mathcal{R}}
\newcommand{\pc}{\mathcal{p}}
\newcommand{\fb}{\mathbf{f}}
\newcommand{\pbf}{\mathbf{p}}
\newcommand{\qbf}{\mathbf{q}}
\newcommand{\kbf}{\mathbf{k}}
\newcommand{\vbf}{\mathbf{v}}
\newcommand{\Qbf}{\mathbf{Q}}
\newcommand{\Kbf}{\mathbf{K}}
\newcommand{\Vbf}{\mathbf{V}}
\newcommand{\Rm}{\mathrm{R}}
\newcommand{\CB}{\mathbf{C}}
\newcommand{\Xbf}{\mathbf{X}}
\newcommand{\PP}{\mathbb{P}}
\newcommand{\alphab}{\mathbb{\alpha}}
\newcommand{\sbtheta}{\mathbf{s}_\theta}
\newcommand{\sbpsi}{\mathbf{s}_\psi}

\newcommand{\RR}{\mathbb{R}}
\newcommand{\NN}{\mathbb{N}}
\newcommand{\Kcal}{\mathcal{K}}
\newcommand{\Rcal}{\mathcal{R}}
\newcommand{\Ncal}{\mathcal{N}}
\newcommand{\Scal}{\mathcal{S}}
\newcommand{\Qcal}{\mathcal{Q}}
\newcommand{\Tcal}{\mathcal{T}}
\newcommand{\Dcal}{\mathcal{D}}

\newcommand{\xtil}{\tilde{x}}

\def\eg{\emph{e.g}.}
\def\ie{\emph{i.e}.}
\def\cf{\emph{c.f}.}
\def\etc{\emph{etc}.}
\def\wrt{w.r.t.}
\def\etal{\emph{et al}.}
\def\ab{\mathbf{a}}
\def\Abf{\mathbf{A}}
\def\xbf{\mathbf{x}}
\def\ybf{\mathbf{y}}
\def\nbf{\mathbf{n}}
\def\bb{\mathbf{b}}
\def\cb{\mathbf{c}}
\def\eb{\mathbf{e}}
\def\Hbf{\mathbf{H}}
\def\hb{\mathbf{h}}
\def\wbf{\mathbf{w}}
\def\Wbf{\mathbf{W}}
\def\zbf{\mathbf{z}}
\def\Zb{\mathbf{Z}}
\def\sbf{\mathbf{s}}
\def\tb{\mathbf{t}}
\def\ub{\mathbf{u}}
\def\Rb{\mathbf{R}}
\def\Lb{\mathbf{L}}
\def\Ib{\mathbf{I}}
\def\fta{f_{\theta}}

\def\rarr{\rightarrow}

\def\Xc{\mathcal{X}}
\def\Yc{\mathcal{Y}}
\def\Zc{\mathcal{Z}}
\def\Pc{\mathcal{P}}
\def\Ic{\mathcal{I}}

\def\wxb{\boldsymbol{w}_x}
\def\wyb{\boldsymbol{w}_y}
\def\muxb{\boldsymbol{\mu}_x}
\def\muyb{\boldsymbol{\mu}_y}
\def\mub{\boldsymbol{\mu}}
\def\pib{\boldsymbol{\pi}}
\def\Sigmab{\boldsymbol{\Sigma}}
\def\xbo{\boldsymbol{x}}
\def\ybo{\boldsymbol{y}}
\def\pb{\boldsymbol{p}}
\def\qb{\boldsymbol{q}}
\def\kb{\boldsymbol{k}}
\def\vb{\boldsymbol{v}}
\def\tb{\boldsymbol{t}}
\def\ub{\boldsymbol{u}}
\def\Xb{\boldsymbol{X}}
\def\Yb{\boldsymbol{Y}}
\def\Vb{\boldsymbol{V}}
\def\Mb{\boldsymbol{M}}
\def\Ab{\boldsymbol{A}}
\def\Bb{\boldsymbol{B}}
\def\Wa{W_a}
\def\Wb{W_b}  %% WARNING Don't be confused with Bold W
\def\Wab{\boldsymbol{W}_a}
\def\Wbb{\boldsymbol{W}_b}
\def\Wxb{\boldsymbol{W}_x}
\def\Wyb{\boldsymbol{W}_y}
\def\Mx{M_x}
\def\My{M_y}
\def\Pb{\boldsymbol{P}}
\def\Wbold{\boldsymbol{W}}
\def\lambdab{\boldsymbol{\lambda}}
\def\dist{\text{d}}
\def\Nc{\mathcal{N}}
\def\epsilonb{\boldsymbol{\epsilon}}
\newcommand{\Mcb}{\boldsymbol{\mathcal{M}}}

\newcommand{\defeq}{\mathrel{\mathop:}=}

\newcommand{\SPD}{\text{SPD}}
\newcommand{\VAR}{\text{VAR}}
\newcommand{\EXP}{\text{Exp}}
\newcommand{\LOG}{\text{Log}}
\newcommand{\EE}{\mathbb{E}}
\newcommand{\x}{\textbf{x}}
\newcommand{\M}{\mathcal{M}}
\newcommand{\Ac}{\mathcal{A}}
\newcommand{\Hc}{\mathcal{H}}
\newtheorem{defn}{Definition}[section]
\newcommand{\tr}{\text{tr}}

\definecolor{cadmiumgreen}{rgb}{0.0, 0.42, 0.24}
\definecolor{calpolypomonagreen}{rgb}{0.12, 0.3, 0.17}
\definecolor{dartmouthgreen}{rgb}{0.05, 0.5, 0.06}
\definecolor{debianred}{rgb}{0.84, 0.04, 0.33}
\definecolor{brightmaroon}{rgb}{0.76, 0.13, 0.28}
\definecolor{cadmiumred}{rgb}{0.89, 0.0, 0.13}
\definecolor{crimson}{rgb}{0.86, 0.08, 0.24}

%% file: Sections/0_abstract.tex
\input{Figures/main_fig}

\begin{abstract}
Recent studies on inverse problems have proposed posterior samplers that leverage the pre-trained diffusion models as powerful priors. 
These attempts have paved the way for using diffusion models in a wide range of inverse problems.
However, the existing methods entail computationally demanding iterative sampling procedures and optimize a separate solution for each measurement, which leads to limited scalability and lack of generalization capability across unseen samples.
To address these limitations, we propose a novel approach, Diffusion prior-based Amortized Variational Inference (DAVI) that solves inverse problems with a diffusion prior from an amortized variational inference perspective.
Specifically, instead of separate measurement-wise optimization, our amortized inference learns a function that directly maps measurements to the implicit posterior distributions of corresponding clean data, enabling a single-step posterior sampling even for unseen measurements.
Extensive experiments on image restoration tasks, \eg, Gaussian deblur, 4$\times$ super-resolution, and box inpainting with two benchmark datasets, demonstrate our approach's superior performance over strong baselines.
Code is available at \url{https://github.com/mlvlab/DAVI}.
  \keywords{Inverse Problems \and Diffusion Models \and Posterior Sampling}
\end{abstract}

%% file: Figures/main_fig.tex
\begin{figure*}[t]
\centering
\includegraphics[width=1\textwidth]{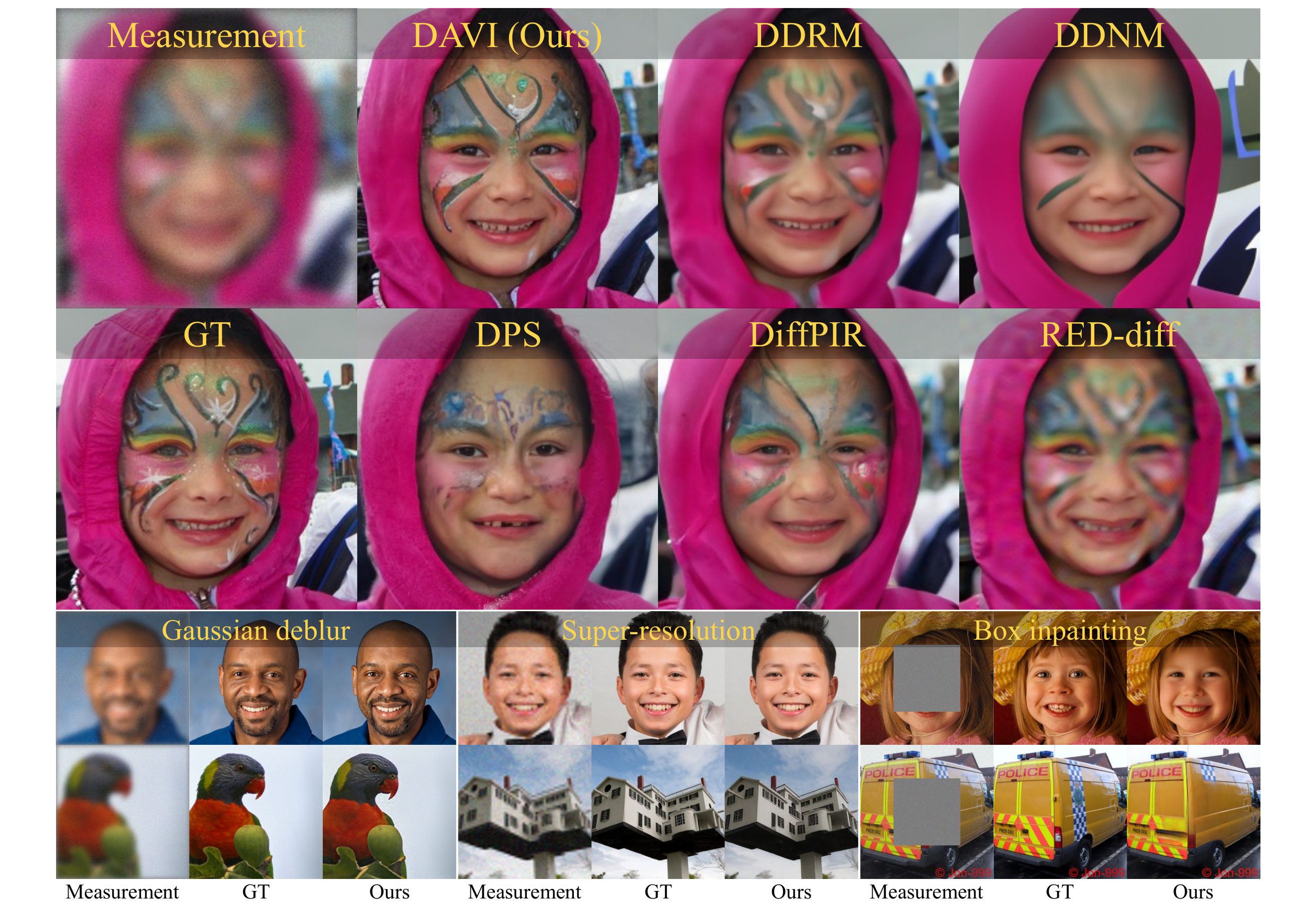}
\captionof{figure}{\textbf{Representative results of Diffusion prior-based Amortized Variational Inference (DAVI)}. 
The top row demonstrates the qualitative comparison between our method and baselines. 
The bottom two rows showcase that DAVI provides robust solutions with fine-grained details across various image restoration tasks, achieved with a \textit{single neural network evaluation}.
}
\label{fig:main_fig}
\end{figure*}

%% file: Sections/1_introduction.tex
\section{Introduction}
\label{sec:intro}
%-------------------------------------------------------------------------
Noisy inverse problems are important and have a wide range of real-world applications such as image restoration~\cite{karras2021alias, zamir2022restormer}, medical imaging~\cite{song2021solving, ozbey2023unsupervised}, and astronomy~\cite{guilloteau2020hyperspectral, takeishi2021physics}. Typically, noisy inverse problems involve estimating the original clean data $\xbf_0$ from a noisy observation $\ybf$.
The forward (measurement) model is formulated as
\begin{equation}
    \ybf = \Hbf {\xbf_0} + \nbf, \quad \nbf \sim \Nc( \mathbf{0},\sigma_{\ybf}^2 \mathbf{I}),
    \label{eq:inverse_problem}
\end{equation}
where a known degradation matrix $\Hbf \in \RR^{{d_\ybf} \times {d_{\xbf_0}}}$ is applied to the clean signal ${\xbf_0} \in \RR^{d_{\xbf_0}}$ with i.i.d white Gaussian noise $\nbf \in \RR^{d_\ybf}$.
Then, the likelihood of the measurement is defined as $p(\ybf|{\xbf_0}) = \Nc(\ybf | \Hbf {\xbf_0}, \sigma_{\ybf}^2 \mathbf{I})$. However, it is challenging to accurately estimate the solution $\xbf_0$ due to its ill-posed nature~\cite{richardson1972bayesian} where there exist multiple solutions $\xbf$ for a measurement $\ybf$, \ie, many-to-one $\xbf \mapsto \ybf$.

%-------------------------------------------------------------------------
Diffusion models have achieved remarkable success across various applications, including image~\cite{kostochastic, yin2024one}, video~\cite{chen2024videocrafter2, wei2024dreamvideo}, 3D~\cite{wang2023prolificdreamer,raj2023dreambooth3d}, and domain-agnostic~\cite{park2024ddmi} generation, encompassing 2D, 3D, and video.
Inspired by these impressive results, recent approaches~\cite{choi2021ilvr, kawar2022denoising, wang2022zero, chung2022improving, chung2023diffusion, song2023pseudoinverse, zhu2023denoising, dou2024diffusion} have leveraged a pre-trained diffusion model~\cite{sohl2015deep, ho2020denoising, song2019generative, karras2022elucidating} as a powerful prior for solving image inverse problems. 
Existing diffusion-based methods alter the reverse sampling process of the pre-trained diffusion model either by approximating the posterior score function via Bayes' rule~\cite{chung2023diffusion,song2023pseudoinverse} or running the reverse process in the decomposed space~\cite{kawar2022denoising,wang2022zero}.
Mardani~\cite{mardani2023variational} proposes a variational approach relying on the diffusion sampling process to approximate the mode of the true posterior distribution.
Despite the promising results of these previous methods, their necessity for the \emph{iterative sampling procedure}
limits their scalability. 
This makes it difficult to deploy the diffusion-based methods on commodity devices for real-time applications.
In addition, \emph{independent optimization} for each sample is suboptimal, leading to poor generalization on unseen samples. 
Arguably, solving the inverse problems for multiple measurements together is more effective if a method seeks a generalizable function beyond one clean sample $\xbf_0$.
The estimated function is readily applicable to unseen samples even without any iterative procedures.

%%-----------------------------------------------------------------------------
Thus, we introduce \textbf{D}iffusion prior-based \textbf{A}mortized \textbf{V}ariational \textbf{I}nference (DAVI) that addresses inverse problems using a diffusion prior within an amortized variational inference framework.
Unlike previous approaches, DAVI learns a function that associates a measurement with the implicit posterior distribution of the corresponding clean data.
This enables efficient single-step posterior sampling for both seen and unseen measurements.
Our method optimizes this function by minimizing the Kullback-Leibler (KL) divergence between the implicit and the true posterior distributions for multiple measurements, employing objectives from variational inference.

Our \textbf{contributions} are summarized as follows: 
\begin{itemize}
\item We propose a novel approach, Diffusion prior-based Amortized Variational Inference (DAVI), which solves inverse problems with a diffusion prior from an amortized variational inference perspective.
\item Our framework enables efficient posterior sampling by a single evaluation of a neural network and generalization for both seen and unseen measurements without any optimization at test time.
\item We propose a novel Perturbed Posterior Bridge that provides intermediary measurements to further enhance the generalization capabilities.
\item Our extensive experiments demonstrate the effectiveness of our proposed method in image restoration tasks.
\end{itemize}

%% file: Sections/2_relatedworks.tex
\section{Related Works}
\label{sec:realatedworks}
% %-------------------------------------------------------------------------
\subsection{Diffusion models for inverse problems}
\label{subsec:rw_inverseproblems_with_diffusion}
Several recent studies~\cite{Chung_2022_CVPR,chung2023diffusion,kawar2022denoising,zhu2023denoising,song2023pseudoinverse,Mokady_2023_CVPR,mardani2023variational,dou2024diffusion} have focused on solving inverse problems using pre-trained diffusion models as priors due to their strong ability to model complex distributions. Specifically, DDRM~\cite{kawar2022denoising} proposes running the diffusion model in a spectral domain through the singular value decomposition (SVD) of the signal space. However, computing SVD can be computationally expensive, especially for high-dimensional signals, and is not feasible for complex degradation operators such as motion blur. While DDNM~\cite{wang2022zero} and $\Pi$GDM~\cite{song2023pseudoinverse} introduce pseudo-inverse guidance during the reverse diffusion process without computing SVD, these methods often struggle with noisy measurements due to the inherent error in estimating the pseudo-inverse with a noisy measurement. 
DPS~\cite{chung2023diffusion} proposes sampling from a diffusion posterior $p(\xbf_0|\ybf)$ by an approximation of the intractable time-dependent likelihood $p(\ybf | \xbf_t)$ using Tweedie's formula.
FPS~\cite{dou2024diffusion} connects Bayesian posterior sampling and Bayesian filtering in diffusion models, where multiple samples share a single measurement.
Like our method, RED-diff~\cite{mardani2023variational} explores the variational perspective of inverse problems by optimizing in pixel space.
However, these methods heavily depend on reverse diffusion trajectories for inference, which inherently requires multiple neural network evaluations. In contrast, our work aims to solve the inverse problem with a single step, which is significantly more efficient.

% %-------------------------------------------------------------------------
\subsection{Amortized variational inference}
\label{subsec:amortized_inference}
Variational inference~\cite{jordan1999introduction,ganguly2023amortized,kingma2013auto} (VI) approximates the posterior with a parameterized variational distribution family $\mathcal{Q}$ and finds the optimal distribution from $\mathcal{Q}$ that minimizes the KL divergence from the true posterior.
Several VI approaches~\cite{steyvers2007probabilistic}, such as the mean-field family, optimize an independent set of variational distributions for each data sample, causing the parameters to scale with the dataset size.
This process, inherently memoryless~\cite{gershman2014amortized}, can pose significant computational inefficiency, particularly for large datasets. On the other hand, Amortized VI (AVI) amortizes the optimization by using a stochastic function, typically a neural network, to represent the entire variational distribution family $\mathcal{Q}$.
AVI memorizes past inferences by optimizing a single neural network across multiple samples rather than optimizing each sample independently.
The optimized network can generalize to unseen samples by using information from previous samples without additional optimization costs.
This allows for computational efficiency compared to per-data optimization. In our work, we adopt amortized VI to address the computationally intensive sampling processes in previous methods, offering a more efficient and scalable solution.

%% file: Sections/3_preliminary.tex
\section{Preliminary}
\label{subsec:rw_diffusion_models}
Diffusion models~\cite{ho2020denoising,song2020score} define stochastic differential equations (SDEs) for the diffusion forward process $\{ {\xbf_t} \}_{t=0}^T$, where $\xbf_t$ is the perturbed data at time $t$.
The data distribution is defined at $t=0$, \ie, $\xbf_0 \sim p(\xbf_0)$ and the prior distribution is achieved at $t=T$, following the standard normal distribution, \ie, $\xbf_T \sim \Nc(\mathbf{0}, \mathbf{I})$.
The forward SDE is defined as
\begin{equation}
    \label{eq:def_forward}
    d\xbf = f_t \xbf_t \; dt + g_t \; d\wbf,
\end{equation}
where $f_t:\RR \rightarrow \RR$ is a drift coefficient, $g_t \in \RR \rightarrow \RR$ is a diffusion coefficient, and $\wbf$ represents a standard Wiener process.
Song et.al~\cite{song2020score} define reverse SDE that achieves the data distribution from the prior distribution using the score function $\nabla_{\xbf_t} \log p(\xbf_t)$ as
\begin{equation}
    \label{eq:def_backward}
    d\xbf = \left[ f_t \xbf_t - g_t^2 \nabla_{\xbf_t} \log p(\xbf_t) \right] dt + g_t \; d\bar{\wbf},
\end{equation}
where $dt$ is an infinitesimal negative timestep and $\bar{\wbf}$ is reverse-time standard Wiener process~\cite{anderson1982reverse}. The time-dependent score function $\nabla_{\xbf_t} \log p(\xbf_t)$ is estimated by a neural network $s_\theta(\xbf_t, t)$, \ie, $\nabla_{\xbf_t} \log p(\xbf_t) \approx s_\theta(\xbf_t, t)$, by minimizing the following denoising score matching objective:
\begin{equation}
    \label{eq:def_scorematching}
    \EE_{p(\xbf_t|\xbf_0), t\sim \mathrm{U}[0,T]}
    \left[ \lambda(t)
    \lVert 
    \mathbf{s}_{\theta}(\xbf_t, t) - \nabla_{\xbf_t} \log p(\xbf_t|\xbf_0)
    \lVert_2^2
    \right],
\end{equation}
where $p(\xbf_t|\xbf_0)$ is a Gaussian transition kernel from time 0 to $t$, and $\lambda(t)$ is a time-dependent weighting function. For Variance Preserving (VP) SDE~\cite{song2020score} or DDPM~\cite{ho2020denoising}, the transition kernel $q(\xbf_t|\xbf_0)$ is defined as $\xbf_t = \sqrt{\alpha_t}\xbf_0 + \sqrt{1-\alpha_t}\epsilon$, $\epsilon \sim \Nc(\mathbf{0}, \mathbf{I})$, where $\alpha_t = \prod_{s=1}^t (1-\beta_t)$ and $\beta_t = g_t^2$.

%% file: Sections/4_method.tex
\input{Figures/framework}
\section{Proposed Method}
\label{sec:method}
We propose a novel approach, Diffusion prior-based Amortized Variational Inference (DAVI), that tackles inverse problems by leveraging a pre-trained diffusion model as a prior distribution from an amortized variational inference perspective~\cite{shu2018amortized,ganguly2023amortized}.
To be specific, our framework learns a neural network that directly maps a measurement $\ybf$ to the implicit posterior distribution $p(\xbf_0|\ybf)$ of the corresponding clean data $\xbf_0$.
This enables efficient posterior sampling via a single evaluation of the neural network and generalization over unseen measurements (\ie, zero-shot inference).
In Section \ref{sec:method_davi} and \ref{sec:method_gradient_derive}, we introduce training objectives that minimize the KL divergence between the implicit distribution and the true posterior distribution based on variational inference.
In Section \ref{sec:method_amortized_inference}, we propose a novel Perturbed Posterior Bridge (PPB) to provide intermediary measurements to further enhance the generalization power of the proposed method.

%-------------------------------------------------------------------------
\subsection{Diffusion prior-based Amortized Variational Inference (DAVI)}
\label{sec:method_davi}
Our framework approximates the posterior distribution $p(\xbf_0|\ybf)$, which is the distribution of the clean sample $x_0$ given the measurement $y$, using an implicit distribution $\Ic_\phi$ parameterized by a neural network. 
Sampling from the approximated distribution is performed by the reparameterization trick with a random Gaussian noise $\zbf \sim \Nc(\mathbf{0},\mathbf{I})$
as follows:
\begin{equation}
    \hat{\xbf} = \Ic_\phi(\ybf + h\cdot \zbf), \ \ \zbf \sim \Nc(\mathbf{0},\mathbf{I}),
    \label{eq:def_implicit_distribution}
\end{equation}
where $\hat{x}$ indicates the sample from the implicit distribution and $h\in \mathbb{R}_{++}$ is a hyperparameter that determines the variance of the noise. 
Variational optimization between our implicit distribution and the true posterior distribution is formulated as 
\begin{equation}
    \phi^\star =
    \arg \min_{\phi} \left[ D_{KL}( q_{\phi}(\xbf_0|\ybf) \parallel p(\xbf_0|\ybf)) \right],
    \label{eq:def_VI}
\end{equation}
where $q_{\phi}(\xbf_0|\ybf)$ and $p(\xbf_0|\ybf)$ are the implicit distribution and the true posterior, respectively.
By the definition of KL-divergence and dropping the constant term `$\log p(\ybf)$', the objective function can be rewritten as
\begin{align}
- \EE_{q_\phi(\xbf_0|\ybf)}
\left[ \log p(\ybf | \xbf_0) \right]
+ D_{KL}( q_{\phi}(\xbf_0|\ybf) \parallel p(\xbf_0)). 
\label{eq:loss_total}
\end{align}
The constant term `$\log p(\ybf)$' can be ignored since it is independent of the parameter $\phi$.
Then, the first term is the likelihood of the measurement $\ybf$, \ie, $\log p(\ybf|\xbf_0)$, which we refer to as a data consistency loss.
The second term, the KL-divergence between the approximated posterior distribution $q_\phi(\xbf_0|\ybf)$ and the prior $p(\xbf_0)$, is represented by the pre-trained diffusion model $p_\theta(\xbf_0)$. We discuss the terms below in further detail. 
\\

%-------------------------------------------------------------------------
\noindent{\textbf{Data consistency loss.}}
The data consistency loss is 
\begin{equation}
    \Lc_{C} =  \EE_{q_\phi(\xbf_0|\ybf)} \left[ \frac{{\lVert \ybf - \Hbf \xbf_0\lVert}^2_2}{2 \sigma_{\ybf}^2} \right],
    \label{eq:loss_consistency}
\end{equation}
since the likelihood $p(\ybf|\xbf_0)$ can be analytically calculated by Eq.~\eqref{eq:inverse_problem}.
Note that the constant term is omitted.
By minimizing the data consistency loss, we incorporate measurement information into the implicit distribution.
\\

%-------------------------------------------------------------------------
\noindent \textbf{Integral KL divergence (IKL).}
Optimizing the KL divergence between approximated posterior distribution and diffusion prior poses a risk of the KL divergence potentially diverging towards infinity, especially when the supports of two distributions are misaligned, leading to unstable and suboptimal results.
To mitigate this issue, Luo et al.~\cite{luo2023diffinstruct} have proposed the Integral KL divergence (IKL). Motivated by the IKL divergence, we upper bound the second term with the integral of KL divergence between $q_\phi(\xbf_t|\ybf)$ and $p_\theta(\xbf_t)$ over $t$, and denote it as $\Lc_{IKL}$. It is formulated as:
\begin{equation}
    \Lc_{IKL} =
    \int_{t=0}^T w(t) D_{KL} (q_{\phi}(\xbf_t|\ybf) \parallel p_\theta(\xbf_t)) dt,
    \label{eq:loss_implicit_posterior}
\end{equation}
where we achieve $\xbf_t$ by the forward SDE in Eq.~\eqref{eq:def_forward} and $w(t)$ is a positive weighting function, notably with $w(0)=1$.
Roughly speaking, the forward SDE can be viewed as a set of Gaussian filters with various bandwidths. It transforms the two distributions into smoother distributions with infinite supports, alleviating the disjoint support problem.
As the perturbation or $t$ increases, the densities overlap more, leading to more robust convergence than the original KL divergence (see Fig.~\ref{fig:IKL}).
This loss can be interpreted as the KL-divergence between two sets of distributions after smoothing by Gaussian kernels with various bandwidths (\ie, the forward diffusion process at various time points).
As $\Lc_{IKL}$ decreases, the samples from the implicit posterior distribution increasingly resemble the cleaner images of the diffusion prior.

%-------------------------------------------------------------------------
\input{Figures/IKL}

\subsection{Score distillation gradient}
\label{sec:method_gradient_derive}
Here, we present the procedure to learn the implicit distribution with the IKL loss. 
The gradient of IKL loss with respect to $\phi$, \ie, $\nabla_\phi \Lc_{IKL}$, can be analytically derived as 
\begin{align}
    \label{eq:loss_gradient}
        \int_{t=0}^T\!\!w(t) \EE_{q_\phi(\xbf_t|\ybf)}\!
        \left[ 
        \nabla_{\xbf_t} \log q_\phi(\xbf_t|\ybf) - \nabla_{\xbf_t} \log p_\theta(\xbf_t)
        \right] 
        \frac{\partial \xbf_t}{\partial \phi} dt, 
\end{align}
where $q_\phi(\xbf_t|\ybf) = \int q(\xbf_t|\xbf_0) \cdot q_\phi(\xbf_0|\ybf) d\xbf_0$.
Here, $q(\xbf_t|\xbf_0)$ is the forward diffusion kernel that is identically defined as $p(\xbf_t|\xbf_0)$ (see supplement Section~\ref{sec:supple_proofs} for detailed derivation).
Eq.~\eqref{eq:loss_gradient} requires the score estimation for both distributions, $p_\theta$ and $q_\phi$, respectively.
We utilize a pre-trained diffusion model $\mathbf{s}_\theta$ to approximate $\nabla_{\xbf_t} \log p(\xbf_t)$; however, a corresponding score function for $q_\phi(\xbf_t|\ybf)$ is not tractable for the implicit distribution.
To address this, we introduce an implicit score function, denoted as $\mathbf{s}_\psi$, and train it to approximate $\nabla_{\xbf_t} \log q_\phi(\xbf_t|\ybf)$.
Then, Eq.~\eqref{eq:loss_gradient} is approximated as
\begin{equation}
    \begin{split}
        \nabla_\phi \Lc_{IKL}
        \approx 
        \EE_{q_\phi(\xbf_t|\ybf), t}
        \left[w(t)
        (\vartriangle \sbf_{\psi, \theta} \cdot \frac{\partial \xbf_t}{\partial \phi})
        \right],
        \label{eq:loss_gradient_scorefunction}
    \end{split}
\end{equation}
where $\vartriangle \sbf_{\psi, \theta} \coloneqq \sbf_{\psi}(\xbf_t,t) - \sbf_{\theta}(\xbf_t,t)$ and $\mathbf{s}_\psi$ indicates the score function of $q_{\phi}$.
Roughly speaking, IKL updates $\phi$ in the direction that minimizes the discrepancy $\vartriangle \sbf_{\psi, \theta}$ between $\mathbf{s}_\psi$ and $\sbtheta$ and 
this eventually minimizes the discrepancy between $q_\phi(\xbf|\ybf)$ and $p_\theta(\xbf)$, as demonstrated in Fig.~\ref{fig:IKL}.

Lastly, we train the score function $\mathbf{s}_\psi$ by the denoising score matching loss described in Eq.~\eqref{eq:def_scorematching} to approximate the score function of $q_\phi$:
\begin{equation}
    \begin{split}
        \Lc_{S} =
        \EE_{q_\phi(\xbf_t|\ybf), t} 
        \left[ 
            \lVert \mathbf{s}_\psi(\xbf_t, t) - \nabla_{\xbf_t} \log q_\phi(\xbf_t|\ybf) \lVert_2^2 
        \right].
        \label{eq:loss_variationalscore}
    \end{split}
\end{equation}

%-------------------------------------------------------------------------
\noindent \textbf{Alternating optimization.}
Since the score of $q_\phi(\xbf_t|\ybf)$ evolves based on the implicit distribution $q_\phi(\xbf_0|\ybf)$ as the optimization progresses, we need to estimate the score function of $q_\phi$ accordingly.
Therefore, we propose an alternating optimization approach for training two parameters $\phi$ and $\psi$ with separate objectives (Eq.~\eqref{eq:loss_total} and Eq.~\eqref{eq:loss_variationalscore}) as described in Algorithm~\ref{alg:alg_davi} and Fig.~\ref{fig:framework}.

\input{Algorithm/main_algorithm}
\input{Algorithm/Inference}

%-------------------------------------------------------------------------
\input{Figures/PPB}

\subsection{Perturbed Posterior Bridge}
\label{sec:method_amortized_inference}
To further improve the generalization power of our framework, we propose a Perturbed Posterior Bridge (PPB), which acts as an intermediary set of trajectories between $\ybf$ and $\xbf$, to facilitate a more guided and effective training of the neural network $\Ic_{\phi}$. 
The sample $\ybf_a$ drawn from PPB is defined as:
\begin{equation}
    \ybf_a = (1-\sigma_a)\ybf + \sigma_a \xbf + h\bar{\sigma}_a\zbf, \ \ \zbf \sim \Nc(\mathbf{0},\mathbf{I}), 
\end{equation}
where $a \in [0, 1]$, $\sigma_a = \frac{\int_{a}^1 \beta_t dt}{\int_{0}^{a}\beta_t dt + \int_{a}^1\beta_t dt}$, and $\beta_t$ is a diffusion noise schedule~\cite{ho2020denoising}. 
The function $\sigma_a$ is defined to ensure the conditions $\sigma_a \in [0, 1]$, $\sigma_0 = 1$, and $\sigma_1 = 0$. This creates a bridge, modulated by $\sigma_a$, with an additional perturbation term $h\bar{\sigma}_a$. The $\bar{\sigma}_a$ determines the deviation from the linear bridge between $\ybf$ and $\xbf$ (see Fig.~\ref{fig:fig_ppb} (B) and (C)). 
Our experiments in Tab.~\ref{tab:ablation} show that the PPB with monotonically increasing $\bar{\sigma}_a$ as approaching $\ybf$ in Fig.~\ref{fig:fig_ppb} (C) is more effective than a constant perturbation Fig.~\ref{fig:fig_ppb} (B). 

Now, we extend the implicit distribution $\Ic_\phi$ with PPB and $a$ as:
\begin{equation}
    \hat{\xbf} = \Ic_{\phi,a}(\ybf_a) = 
    \begin{cases}
    \Ic_{\phi,0}(\xbf), & \text{if} \ a=0 \\
    \Ic_{\phi,a}((1 - \sigma_a)\ybf + \sigma_a \xbf + \left ( \sigma_a (1-\sigma_a) + h\bar{\sigma}_a \right )\zbf), & \text{elif} \ 0 < a < 1 \\ 
    \Ic_{\phi,1}(\ybf + h\zbf), & \text{otherwise} \\ 
    \end{cases} \\
    \label{eq:PPB_implicit_distribution}
\end{equation}
where $a$ is encoded using the sinusoidal positional embedding~\cite{vaswani2017attention} and fed into each block of the neural network $\Ic_{\phi,a}$. 
Note that when $a=1$, Eq.~\ref{eq:PPB_implicit_distribution} is reduced to the implicit distribution defined in Eq.~\ref{eq:def_implicit_distribution}. 
This approach allows us to optimize the implicit distribution with more diverse samples drawn from PPB between $\ybf$ and $\xbf$. 
We observed it effectively enhances the generalization to unseen measurements. 
During the training stage, we randomly sample $a$ from pre-defined distribution $p(a)$, \ie, $a \sim p(a)$, and the overall training procedure is described in Alg.~\ref{alg:alg_davi}.
\\

\noindent \textbf{Inference.} Our single step inference is described in Alg.~\ref{alg:inference}. Our framework generalizes well on unseen measurements $\ybf$.
We sample a random noise $\zbf \sim \Nc (\mathbf{0},\mathbf{I})$, and obtain the posterior sample via a single evaluation of a neural network as $\hat{\xbf} = \Ic_{\phi,1}(\ybf+h\zbf)$. 

%% file: Figures/framework.tex
\begin{figure*}[t]
    \centering
    \includegraphics[width=1\textwidth]{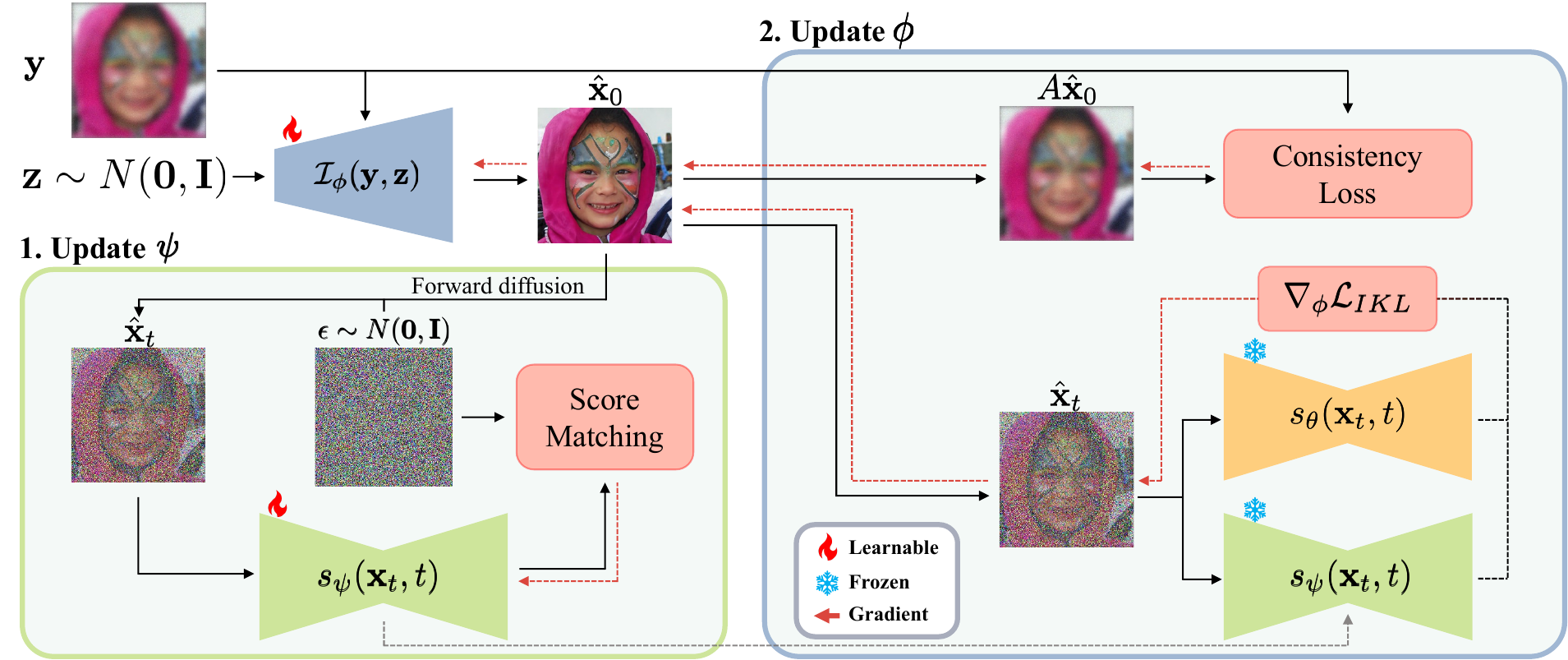}
    \caption{\textbf{Illustration of Diffusion prior-based Amortized Variational Inference (DAVI).} Our proposed method employs an alternative optimization procedure between the score function $\mathbf{s}_\psi$ and the neural network $\Ic_\phi$ to minimize the KL divergence between the implicit posterior distribution $q_\phi(\xbf_0|\ybf)$ and the true posterior distribution $p(\xbf_0|\ybf)$, where the true posterior distribution is approximated by the likelihood $p(\ybf|\xbf_0)$ and the diffusion prior $\mathbf{s}_\theta$.}
    \label{fig:framework}
\end{figure*}

%% file: Figures/IKL.tex
\begin{figure}[t]
    \centering
    \includegraphics[width=1\textwidth]{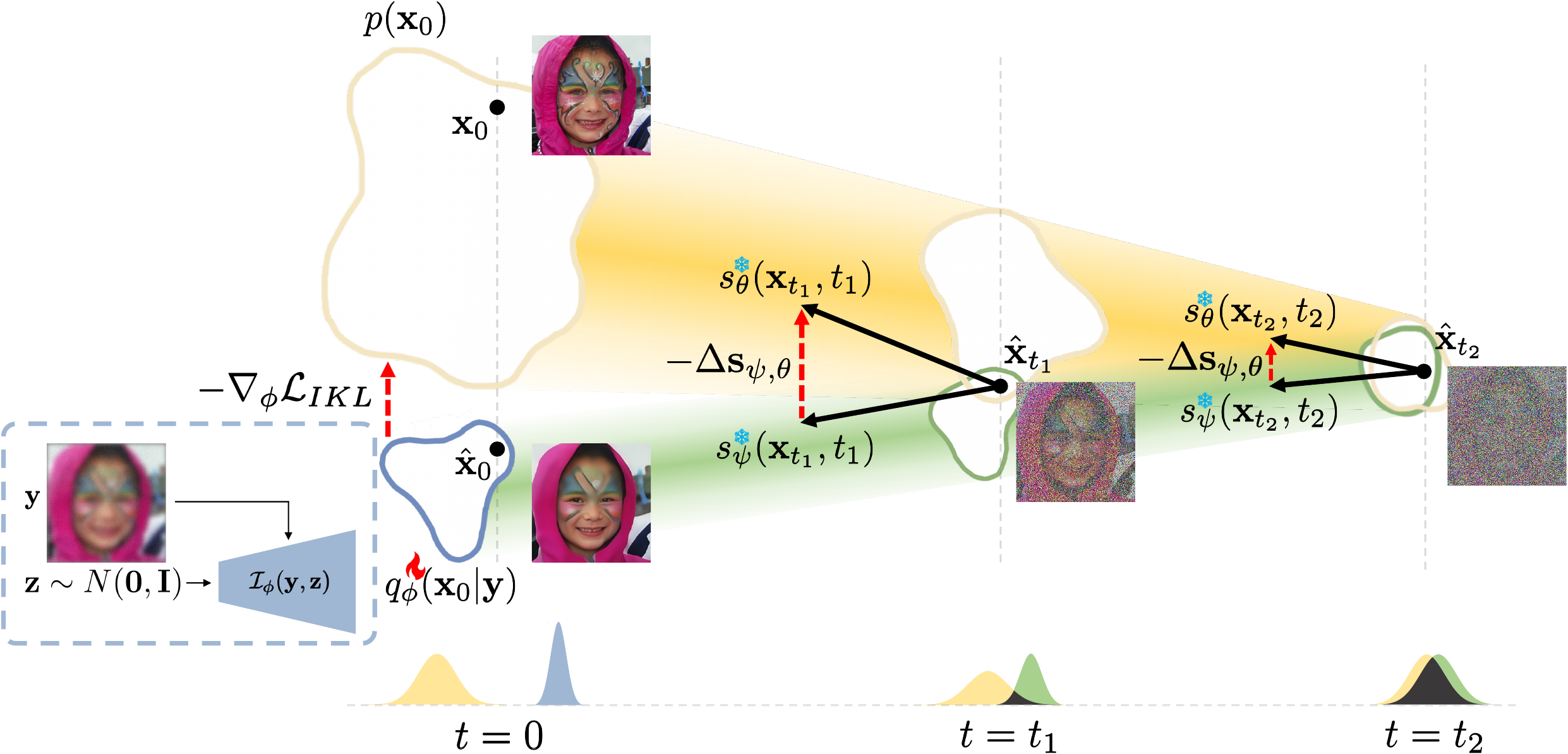}
    \caption{\textbf{Visualization of integral KL divergence.} IKL loss perturbs the implicit distribution into a smoother distribution using forward SDE to alleviate the disjoint support problem (see $t=t_1, t=t_2$, and $t_1 < t_2$). The gradient of IKL loss, $\nabla_\phi \Lc_{IKL}$, updates the parameters of the implicit distribution in a direction that minimizes the $\vartriangle \sbf_{\psi, \theta}$, which leads to minimizing the discrepancy between $q_\phi(\xbf_0|\ybf)$ and $p(\xbf_0)$.
    }
    \label{fig:IKL}
\end{figure}

%% file: Algorithm/main_algorithm.tex
\begin{algorithm}[!t]
    \caption{Training}
    \label{alg:alg_davi}
    \begin{algorithmic}[1]
    \Require $\Hbf, K, T, \mathcal{I}_{\phi,a}, \mathbf{s}_\psi, \mathbf{s}_\theta, w, \gamma, h$
    \For {$k=1, ..., K$}
        \State $\xbf_0 \sim p(\xbf_0)$
        \State $\ybf \sim p(\ybf|\xbf_0)$
        \State $\ybf_a = (1-\sigma_a)\ybf + \sigma_a \xbf_0 + h\bar{\sigma}_a \zbf, \; \text{where} \; a \sim p(a), \ z \sim \Nc( \mathbf{0},\mathbf{I})$
        \State $\hat{\xbf}_0 = \mathcal{I}_{\phi,a}(\ybf_a)$
            \\
            \State $t \sim \mathcal{U}(0,T)$
            \State Draw $\hat{\xbf}_t \sim q(\xbf_t|\xbf_0) \; \text{using Forward SDE (Eq.~\eqref{eq:def_forward}})$
            \\
            \State $\Lc_{S} = \lVert{\mathbf{s}_\psi(\xbf_t,t) - \nabla_{\xbf_t} \log q_\phi(\xbf_t|\ybf)}\lVert_2^2 \;\text{ (Eq.~\eqref{eq:loss_variationalscore})}$
            \State Update $\psi$ with $\nabla_\psi \Lc_{S}$
            \\
            \State $\Lc_{C} = \gamma \lVert{\ybf - \Hbf \hat{\xbf}_0}\lVert_2^2| \; \text{(Eq.~\eqref{eq:loss_consistency})}$
            \State $\nabla_\phi \Lc_{IKL} = w(t) (\mathbf{s}_\psi (\xbf_t, t) - \mathbf{s}_\theta(\xbf_t, t)) \frac{\partial \xbf_t}{\partial \phi} \; \text{(Eq.~\eqref{eq:loss_gradient_scorefunction})}$
            \State Update $\phi$ with $\nabla_\phi (\Lc_{C} + \Lc_{IKL})$
        \EndFor \\
        \Return $\mathcal{I}_{\phi,a}$
    \end{algorithmic}
\end{algorithm}

%% file: Algorithm/Inference.tex
\begin{algorithm}[!t]
    \caption{Single-step Inference}
    \label{alg:inference}
    \begin{algorithmic}[1]
    \Require \sj{$\ybf, \mathcal{I}_{\phi,a}$}
    \State \sj{$\hat{\xbf}_0 = \mathcal{I}_{\phi,1}(\ybf+h\zbf), \ \ \zbf \sim \Nc( \mathbf{0},\mathbf{I})$}
    \\
    \Return $\hat{\xbf}_0$
    \end{algorithmic}
\end{algorithm}

%% file: Figures/PPB.tex
\begin{figure*}[t]
\centering
\includegraphics[width=1\textwidth]{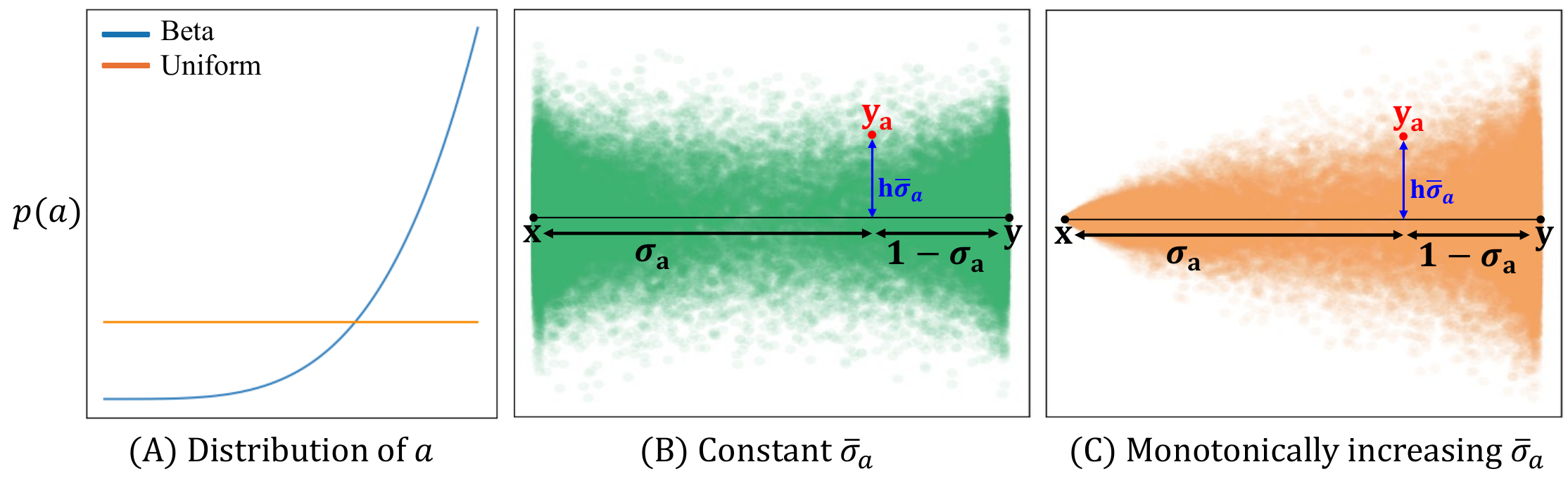}
\captionof{figure}{\textbf{Perturbed Posterior Bridge (PPB).} (A) shows sampling distributions of $a$. (B) and (C) illustrate the PPB between two 1D samples, \eg, $\xbf=0$ and $\ybf=1$, with different perturbation schedule $\bar{\sigma}_a$. We plot the perturbation along the $y$ axis.}
\label{fig:fig_ppb}
\end{figure*}

%% file: Sections/5_experiments.tex
\section{Experiments}
\label{sec:experiments}

\subsection{Experimental Setups}
\label{sec:experiments_setup}
\noindent{\textbf{Baselines.}}
To demonstrate the efficacy of our method for solving noisy inverse problems, we compare DAVI with state-of-the-art methods such as DDRM~\cite{kawar2022denoising}, DDNM$^+$~\cite{wang2022zero}, DPS~\cite{chung2023diffusion}, $\Pi$GDM~\cite{song2023pseudoinverse}, DiffPIR~\cite{zhu2023denoising}, and RED-diff~\cite{mardani2023variational}.
The comparisons are conducted on the FFHQ~\cite{karras2019style} 1K and ImageNet~\cite{ILSVRC15} 1K validation datasets at a resolution of 256$\times$256.
\\
\noindent{\textbf{Tasks.}}
We focus on three challenging image restoration tasks: \textit{Gaussian deblurring}, 4$\times$\textit{super-resolution}, and \textit{box inpainting} with 128$\times$128 size of the mask.
Specifically, we employ a Gaussian blur kernel with 61$\times$61 size, a standard deviation of 3.0 for deblurring, and an average-pooling operation for super-resolution.
Additionally, Gaussian noise with a standard deviation ($\sigma_\ybf$) of 0.05 is introduced to the images to simulate measurement noise unless stated otherwise.
In Section~\ref{sec:supple_deno_color} of the supplement, we provide additional experiments such as \textit{denoising} and \textit{colorization} tasks, and adopt Poisson noise results.
\\
\noindent{\textbf{Evaluation metrics.}}
For evaluation, we focus on two aspects: 1) \textit{consistency} with the original measurement and 2) the \textit{realism} of the restored images. We employ the Peak Signal-to-Noise Ratio (PSNR) to assess measurement fidelity; however, it is noteworthy that PSNR prefers blurry images~\cite{zhang2018unreasonable}.
To address this, we complement it with the Learned Perceptual Image Patch Similarity (LPIPS) score~\cite{zhang2018unreasonable}, which measures structural fidelity.
To evaluate the realism of restored images, we employ the Frechet Inception Distance (FID)~\cite{heusel2017gans}, which better aligns with human visual perception. We also report the number of function evaluations (NFEs) to compare the method's efficiency.
\\
\noindent{\textbf{Implementation details.}}
All methods utilize a pre-trained unconditional diffusion model $\mathbf{s}_\theta$ used in \cite{chung2023diffusion}, for the prior $p_\theta(\xbf_0)$. We employ the same architecture for the implicit distribution $\Ic_{\phi}$ and the implicit score function $\mathbf{s}_\psi$, and initialize $\Ic_{\phi}$ and $\mathbf{s}_\psi$ with the same pre-trained diffusion model for better convergence. In the optimization stage, we use the dataset that trained the pre-trained diffusion model $\mathbf{s}_\theta$ to obtain clean and degraded pairs. We set $p(a)$ as a beta distribution and $\bar{\sigma}_a$ as a monotonically increasing function with $a$. Note that we measure the performance on a \textit{distinct} validation set. For further details, refer to the supplement Section~\ref{sec:implementation}

\input{Tables/ffhq}
\input{Tables/imagenet}
\input{Tables/random_noise_imagenet}

\subsection{Quantitative Results}
\label{sec:experiments_quantity}
We demonstrate the quantitative results on FFHQ and ImageNet in Tab.~\ref{tab:ffhq} and \ref{tab:imagenet}, respectively.
The results clearly show that our proposed method, DAVI, surpasses the leading works in all scenarios, especially in FID and LPIPS, even in a \textit{single-step}.
For example, on the super-resolution task, DAVI achieves a remarkable improvement in FID score, with a gain of 12.33 points on FFHQ and 7.36 points on ImageNest compared to DPS, which records the best FID score among the baselines.
Moreover, on the box inpainting task, DAVI enhances the FID score by 16.96 points on FFHQ and 5.34 points on ImageNet compared to DiffPIR.
On the Gaussian deblurring task, we improve the LPIPS score by 0.044 points on FFHQ and 0.041 points on ImageNet compared to DDRM.
DAVI also exhibits comparable or superior PSNR compared to the other methods. 
In short, these quantitative analyses highlight that DAVI is more effective than the baselines in achieving favorable solutions for inverse problems regarding consistency and realism.
\\
\noindent{\textbf{Robustness to unknown noise scale $\sigma_\ybf$.}}
In our main experiments, we initially set the measurement noise to $\sigma_{\ybf} = 0.05$. However, to account for the variability of noise levels encountered in real-world applications, we extend our evaluation by sampling $\sigma_{\ybf}$ from a uniform distribution $\mathcal{U}(0.01, 0.1)$ for Gaussian deblurring task. This approach aims to emulate more challenging scenarios.
Tab.~\ref{tab:random_noise_imagenet} demonstrates the robustness of DAVI across various scales of $\sigma_\mathbf{y}$ for noisy inverse problems without knowing the noise scale in the measurements.
On the other hand, $\Pi$GDM, DDRM, and DDNM$^+$ show significant performance degradations compared to Tab.~\ref{tab:imagenet} since these approaches require the noise scale $\sigma_\ybf$ for their algorithms and leverage this information to compensate for error in estimating pseudo-inverse or SVD.

\input{Figures/qualitative_result}

\subsection{Qualitative Results}
\label{sec:experiments_quality}
Fig.~\ref{fig:qualitative_comparison} shows qualitative comparisons of our model (DAVI) against baseline methods, selected based on their performance in Tab.~\ref{tab:ffhq} and \ref{tab:imagenet}.
The results reveal that DAVI outperforms the baseline methods in terms of both measurement fidelity and realism. 
For instance, the microphone in the 2nd row of Fig.~\ref{fig:qualitative_comparison} gets blurry in DDRM, DPS, and RED-diff. 
In contrast, DPS and DiffPIR generate realistic and vivid images similar to ours; however, they often remove or change the details in the measurement.
For example, in the sock and crow image (1st and 3rd row, respectively), DPS and DiffPIR struggle to capture small details in measurements, like eyes or green spots.
For the box inpainting task, DAVI provides the most realistic results in the masked areas.
This analysis aligns with the quantitative results, where DPS and DiffPIR, despite ranking high in FID scores for realism, score low in PSNR for consistency.
On the other hand, DAVI generates realistic details in a \textit{single step} and maintains consistency with the measurements.
Please refer to the supplement Section~\ref{sec:supple_qual} for further qualitative results.

\input{Tables/speed_inference}
\input{Tables/compare_setting}

\subsection{Analysis}
\noindent{\textbf{Inference speed.}}
Our efficient single-step posterior sampling achieves an inference time of 0.04 seconds per image, outperforming the fastest baseline DDRM by 49.7$\%$ as demonstrated in Tab.~\ref{tab:speed}. This clearly showcases the efficiency of our framework compared to the baselines.
Moreover, DAVI demonstrates exceptional FID scores, indicating its high-quality performance in solving inverse problems.
In contrast, other methodologies require iterative optimization for each measurement, leading to increased inference times as the number of inferences grows.
\\
\noindent{\textbf{Ablation study.}}
Here, we present an ablation study to investigate the contributions of the proposed components in DAVI. 
Specifically, we examine the improvement by 1) integral KL divergence (IKL) and 2) Perturbed Posterior Bridge (PBB). 
The configurations and results are outlined in Tab.~\ref{tab:ablation}.
We first assess the role of IKL in enhancing the realism of the restored images by minimizing the discrepancy between $q_\phi(\xbf_0|\ybf)$ and $p(\xbf)$.
Comparing configurations A (without IKL, setting $T=0$ in Alg.~\ref{alg:alg_davi}) and B (with IKL), we observe a significant FID score improvement in B by 13.08 points, underscoring IKL's effectiveness.
From configuration C to F, we evaluate the importance of the Perturbed Posterior Bridge, including the perturbation schedule $\bar{\sigma}_a$ and sampling distribution $p(a)$.
Configuration C transforms the implicit distribution to a non-variational one by eliminating perturbation ($h=0$). This transforms the PBB into a line segment between $\xbf_0$ and $\ybf$.
This results in a notable degradation in image quality metrics (FID) compared to B. Conversely, introducing perturbations ($h=0.01$) in D improves the LPIPS and FID, demonstrating the critical role of the PBB in enhancing high-quality image restoration. 
Notably, configuration F shows the most superior performance, with the FID improvement of 17.49 points compared to D, underscoring the significance of $\bar{\sigma}_a$ and $p(a)$ for optimizing the implicit distribution. This suggests that making the PPB generate intermediary measurements closer to $\ybf$ is more effective.

In supplement Sections~\ref{sec:supple_daviG} and \ref{sec:supple_zeroshot}, we evaluate zero-shot capabilities on out-of-distribution data and an unknown degradation model, and train with generated data. In supplement Section~\ref{sec:supple_i2i}, we compare our method with image-to-image translation methods like I$^2$SB~\cite{liu20232} and CDDB~\cite{chung2024direct}.
For more details, please refer to the supplementary materials.

%% file: Tables/ffhq.tex
\begin{table*}[t]
\centering
\caption{\textbf{Comparative results for noisy inverse problems (Gaussian deblur, super-resolution, and box inpainting with centered 128x128 mask) on the FFHQ 256$\times$256}. The highest performance is highlighted in bold, while the second-highest is underlined.}
\resizebox{0.99\linewidth}{!}{
\begin{tabular}{l|c|ccccccccccc} 
\hline
       &             & \multicolumn{3}{c}{Gaussian deblur}                     &  & \multicolumn{3}{c}{4$\times$ Super-resolution} &  & \multicolumn{3}{c}{Box Inpainting}  \\
Method      & NFE $\downarrow$   & PSNR $\uparrow$ & LPIPS $\downarrow$ & FID $\downarrow$ &  & PSNR $\uparrow$ & LPIPS $\downarrow$ & FID $\downarrow$                   &  & PSNR $\uparrow$ & LPIPS $\downarrow$ & FID $\downarrow$                              \\ 
\hline
\hline
DDRM~\cite{kawar2022denoising}   &     20       &    \underline{26.26} &      \underline{0.269}         &   56.82          &  &     28.09 &  0.228  &    48.19      &                       & 22.27 & \underline{0.177} & 37.62                             \\
DDNM$^+$\cite{wang2022zero}       &    100    &    24.19    &       0.348        &    91.48        &  &   \underline{28.17}         &   0.244      &   59.09                    &  &  23.79 &  0.218 & 48.82                              \\
DPS~\cite{chung2023diffusion}        &   1000    &    21.88    &        0.315       &    34.47        &  &      25.55      &    0.255      &     \underline{36.29}                  &  & 21.94 & 0.302 & 42.45                              \\
$\Pi$GDM~\cite{song2023pseudoinverse}        &   100    &    23.05   & 0.320 &  52.72    &  &   27.73    &  \underline{0.225}  &  49.86                   &  & 23.04 & 0.220 & 32.99                             \\
DiffPIR~\cite{zhu2023denoising}  &    100    & 24.41           & 0.299              & \underline{33.91}            &  & 25.32          & 0.341          & 40.35                      &  & 23.97 & 0.195 & \underline{31.27}                            \\ 
RED-diff~\cite{mardani2023variational}  &     1000      & \textbf{26.44}           & 0.324             & 46.55            &  & 26.75          & 0.379          & 92.82                      &  & \underline{24.16} & 0.216 & 35.80       \\
\hline
\textbf{DAVI (Ours)}         &   \textbf{1}    &    25.46    &   \textbf{0.225}            &   \textbf{29.92}         &   &   \textbf{28.23}          &  \textbf{0.171}       &  \textbf{23.96}                       &  & \textbf{26.25} & \textbf{0.112}  &  \textbf{14.31}                              \\
\hline
\end{tabular}
}
\label{tab:ffhq}
\end{table*}

%% file: Tables/imagenet.tex
\begin{table*}[t]
\centering
\caption{\textbf{Comparative results for noisy inverse problems (Gaussian deblur, super-resolution, and box inpainting with centered 128x128 mask) on the ImageNet 256$\times$256}. The highest performance is highlighted in bold, while the second-highest is underlined.}
\resizebox{0.99\linewidth}{!}{
\begin{tabular}{l|c|ccccccccccc} 
\hline
       &             & \multicolumn{3}{c}{Gaussian deblur}                     &  & \multicolumn{3}{c}{4$\times$ Super-resolution} &  & \multicolumn{3}{c}{Box Inpainting}  \\
Method      & NFE $\downarrow$       & PSNR $\uparrow$ & LPIPS $\downarrow$ & FID $\downarrow$ &  & PSNR $\uparrow$ & LPIPS $\downarrow$ & FID $\downarrow$                   &  & PSNR $\uparrow$ & LPIPS $\downarrow$ & FID $\downarrow$                              \\ 
\hline
\hline
DDRM~\cite{kawar2022denoising}   &     20       &    \textbf{23.96} &     \underline{0.384}        &   69.59         &  &     25.83 & 0.300  &   48.47       &                       & 18.10 & 0.260 & 75.32                             \\
DDNM$^+$~\cite{wang2022zero}       &    100    &    22.16    &       0.486        &    120.16        &  &   \underline{26.37}    &       \underline{0.278}        &    46.05                      &  &  20.17 &  0.285 & 92.34                             \\
DPS~\cite{chung2023diffusion}        &   1000    &    19.86    &        0.444      &    64.42        &  &      24.92      &    0.311      &     \underline{43.63}                   &  & 18.87 & 0.391 & 72.48                              \\
$\Pi$GDM~\cite{song2023pseudoinverse}        &   100    &   21.72   &     0.443     &   81.32    &  &    25.53    &   0.321    &   60.61                 &  & 18.83 & \underline{0.233} & 71.42                             \\
DiffPIR~\cite{zhu2023denoising}  &    100       & 22.10           & 0.400              & \textbf{62.48}            &  & 23.08          & 0.385          & 57.39                      &  & 20.26 & 0.259 & \underline{68.38}                            \\ 
RED-diff~\cite{mardani2023variational}  &     1000      & 23.66           & 0.448             & 100.80            &  & 24.82          & 0.406          & 84.72                      &  & \underline{20.26} & 0.276 & 74.38 \\
\hline
\textbf{DAVI (Ours)}         &   \textbf{1}    &    {\underline{23.73}}    &    \textbf{0.343}           &   \underline{63.29}        &   & \textbf{26.58}            &   \textbf{0.242}         &   \textbf{36.27}                      &  & \textbf{21.96} &  \textbf{0.207}  &  \textbf{63.04}                               \\
\hline
\end{tabular}
}
\label{tab:imagenet}
\end{table*}

%% file: Tables/random_noise_imagenet.tex
\begin{table}[t]
\centering
\caption{\textbf{Robustness to the unknown noise scale $\sigma_\ybf$.} Comparative results for Gaussian deblurring on the ImageNet 256$\times$256.
The best results are highlighted in bold, while the second-highest are underlined. "Required" indicates that the noise scale $\sigma_\ybf$ is utilized for the method's operation.}
\resizebox{1\columnwidth}{!}{
\begin{tabular}{l|cccccc|cccccc} 
\hline
Method & \textbf{DAVI} (Ours) & & RED-diff~\cite{mardani2023variational} & & DPS~\cite{chung2023diffusion}  & & $\Pi$GDM~\cite{song2023pseudoinverse}   & & DDRM~\cite{kawar2022denoising} & & DDNM$^+$~\cite{wang2022zero} \\
\hline
Noise scale $\sigma_\ybf$ &  - & & - & & - & & Required & & Required & & Required \\
\hline
PSNR $\uparrow$ & \textbf{23.47} & & \underline{23.25} & & 19.81 & & 21.80 & & 16.09 & & 19.25 \\
LPIPS $\downarrow$ & \textbf{0.347} & & 0.470 &  & 0.447 & & \underline{0.446} & & 0.595 & & 0.543 \\
FID $\downarrow$ & \textbf{63.37}  & & 111.08 & & \underline{64.00} & & 82.98 & & 157.38 & & 147.30 \\
\hline
\end{tabular}
}
\label{tab:random_noise_imagenet}
\end{table}

%% file: Figures/qualitative_result.tex
\begin{figure*}[!ht]
    \centering
    \includegraphics[width=0.95\textwidth]{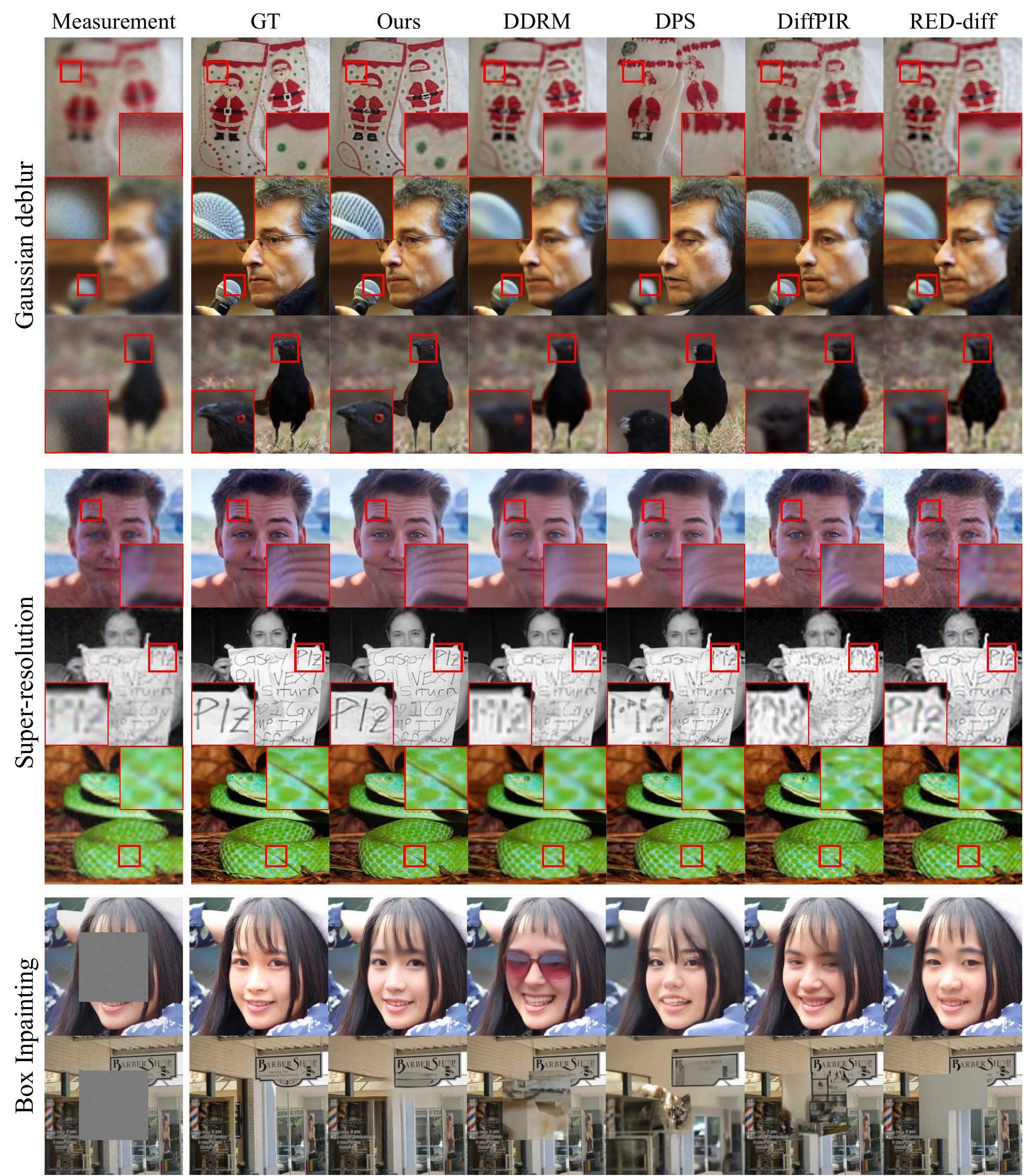}
    \caption{\textbf{Qualitative comparison.} DAVI shows the most vivid and realistic solutions while maintaining intricate details of measurement, in contrast to baselines, which struggle to satisfy both aspects, as highlighted in red boxes.}
    \label{fig:qualitative_comparison}
\end{figure*}

%% file: Tables/speed_inference.tex
\begin{table}[t]
\centering
\caption{\textbf{Inference speed (sec/image).} Wall-clock time on a single TITAN RTX GPU for 4$\times$ Super-resolution on FFHQ 256$\times$256. The highest performance is highlighted in bold, followed by the second-best in underline.}
\resizebox{0.95\columnwidth}{!}{
\begin{tabular}{l|ccccccccccccc} 
\hline
Method & \textbf{DAVI} & & DDRM & & DDNM$^+$ & & DPS  & & $\Pi$GDM & & DiffPIR & & RED-diff  \\
     &  (Ours) & &  \cite{kawar2022denoising}&  & \cite{wang2022zero}&  & \cite{chung2023diffusion}&   & \cite{song2023pseudoinverse}&  & \cite{zhu2023denoising}&  & \cite{mardani2023variational} \\
\hline
time(sec/img) $\downarrow$ & \textbf{0.04}&  & \underline{1.7}&  & 4.2&  & 77.7&  & 8.3&  & 4.3&  & 41.1 \\
\hline
FID $\downarrow$ &  \textbf{23.96} &  & 48.19 & & 59.09 & & \underline{36.29} &  & 49.86 & & 40.35 &  & 92.82 \\
\hline
\end{tabular}
}
\label{tab:speed}
\end{table}

%% file: Tables/compare_setting.tex
\begin{table}[t]
\centering
\caption{\textbf{Ablation study.} We conduct an ablation study on the Gaussian deblurring task on the FFHQ dataset. \textit{Const.} indicates the constant perturbation scale, and \textit{M.I} indicates a monotonically increasing schedule for $a$. Details are in the supplement.}
\resizebox{0.95\linewidth}{!}{
\begin{tabular}{l|c|c|ccc|ccc} 
\hline
Config. & IKL & PPB & $\bar{\sigma}_a$ & $h$ & $p(a)$ & PSNR $\uparrow$ & LPIPS $\downarrow$ & FID $\downarrow$ \\ 
\hline
\hline
(A) & \xmark &  \xmark & -  & - &  - &  21.47  &  0.443 & 50.69        \\
(B) & \checkmark & \xmark &   - & - &  -   & 24.70   &  0.236 & 37.61    \\
\hline
(C) & \checkmark & \checkmark &   Const. &  0 &  Uniform   &  \textbf{26.71} &  0.247 & 49.37    \\
(D) & \checkmark & \checkmark &  Const. & 0.01 &   Uniform  &  \underline{26.17}   & 0.239 & 47.41   \\
(E) & \checkmark & \checkmark &  Const. & 0.01 &   Beta  & 25.99  & \underline{0.232}  & \underline{34.34}    \\
(F) & \checkmark & \checkmark &  M.I  & 0.01 &  Beta  & 25.46   & \textbf{0.225} & \textbf{29.92}     \\
\hline
\end{tabular}
\label{tab:ablation}
}
\end{table}

%% file: Sections/6_conclusion.tex
\section{Conclusion}
\label{sec:conclusion}
In our paper, we propose Diffusion prior-based Amortized Variational Inference (DAVI), a novel framework for solving noisy inverse problems based on amortized variational inference.
DAVI optimizes an implicit distribution parameterized by a neural network by minimizing the KL divergence between the implicit distribution and the true posterior, enabling efficient single-step posterior sampling from implicit distribution by eliminating measurement-wise optimization.
We introduce a novel Perturbed Posterior Bridge (PPB) to enhance the generalizability of our implicit distribution.
Extensive experiments on image restoration tasks show that DAVI outperforms previous state-of-the-art diffusion-based methods.
Lastly, our work uses the human dataset, which may raise ethical concerns that the restored images might resemble real individuals. We are fully aware of such ethical concerns and emphasize the critical need for ethical mindfulness and the development of robust strategies to address potential concerns.

%% file: Section_supple/A_derivations.tex
In Section~\ref{sec:supple_proofs}, we provide a derivation for the analytical formulation of the integral KL divergence. We also examine alternative posterior distributions, facilitating a comparative evaluation of our method against these alternatives, such as RED-diff~\cite{mardani2023variational}.
In Section~\ref{sec:supple_addtional}, we extend the evaluation with additional analyses that underscore the versatility and zero-shot generalization capability of our proposed method. This section also encompasses a comparative study with current image-to-image translation models to broaden the scope of our comparative analysis and highlight the distinct advantages of our method.
In Section~\ref{sec:supple_details}, we provide the experimental setup and details of re-implementing the baseline methods.
In Section~\ref{sec:supple_related_works}, we introduce additional related works utilizing score distillation of diffusion models.
Lastly, Section~\ref{sec:supple_qual} showcases additional qualitative results.

\section{Derivations}
\label{sec:supple_proofs}
\subsection{Derivation of the gradient of IKL divergence}
To make this paper self-contained, we calculate the gradient of the integral KL divergence for our formulation, mostly following the derivation in ~\cite{luo2023diffinstruct}. 
The density $q_\phi(\xbf_t| \ybf)$ contains the parameter $\phi$ implicitly since $q_\phi(\xbf_t| \ybf)$ is initialized with $q_\phi(\xbf_0| \ybf)$ which is generated by our model $\mathcal{I}_\phi$. 
Then the gradient of $\Lc_{IKL}$ with respect to the implicit distribution parameter $\phi$ is given as
\begin{align}
    \nabla_\phi \Lc_{IKL} & =
    \frac{d}{d\phi}\int_{t=0}^T w(t) D_{KL} (q_{\phi}(\xbf_t|\ybf) \parallel p(\xbf_t)) dt 
    \\
    & = \frac{d}{d\phi}\int_{t=0}^T w(t) \mathbb{E}_{q_{\phi}(\xbf_t|\ybf)}\left[ \log q_{\phi}(\xbf_t|\ybf) - \log p(\xbf_t) \right] dt \\
    & = \int_{t=0}^T w(t) \mathbb{E}_{q_{\phi}(\xbf_t|\ybf)}\frac{d}{d\phi}\left[ \log q_{\phi}(\xbf_t|\ybf) - \log p(\xbf_t) \right] dt \label{eq:15} \\
    & = \int_{t=0}^T w(t) \mathbb{E}_{q_{\phi}(\xbf_t|\ybf)} \left[ \nabla_{\xbf_t} \log q_{\phi}(\xbf_t|\ybf) - \nabla_{\xbf_t} \log p(\xbf_t) \right] \frac{d \xbf_t}{d\phi} dt  \\ 
    & \ \ \ \ \ + \int_{t=0}^T w(t) \mathbb{E}_{q_{\phi}(\xbf_t|\ybf)} \left[ \frac{d}{d\phi}\log q_\phi(\xbf_t|\ybf) \right]  dt \label{eq:16} ,
\end{align}
where the chain rule and gradient product rule are applied to Eq.~\eqref{eq:15}. The last term in Eq.~\eqref{eq:16} is equal to zero as

\begin{align}
    \int_{t=0}^T w(t) & \mathbb{E}_{q_{\phi}(\xbf_t|\ybf)} \left[ \frac{d}{d\phi}\log q_\phi(\xbf_t|\ybf) \right]  dt \\  
    & = \int_{t=0}^T w(t) \mathbb{E}_{q_{\phi}(\xbf_t|\ybf)} \left[ \frac{1}{q_\phi(\xbf_t|\ybf)} \frac{d}{d\phi}q_\phi(\xbf_t|\ybf) \right] dt \\
    & = \int_{t=0}^T w(t) \int  q_\phi(\xbf_t|\ybf) \frac{1}{q_\phi(\xbf_t|\ybf)} \frac{d}{d\phi} q_\phi(\xbf_t|\ybf) d\xbf_t dt  \\
    & = \int_{t=0}^T w(t) \int \frac{d}{d\phi} q_\phi(\xbf_t|\ybf) d\xbf_t dt  \\
    & = \int_{t=0}^T w(t) \frac{d}{d\phi} \int q_\phi(\xbf_t|\ybf) d\xbf_t dt  \\
    & = \int_{t=0}^T w(t) \frac{d}{d\phi} \textbf{1} dt  \\
    & = 0.
\end{align}
Therefore, the gradient of $\Lc_{IKL}$ with respect to the implicit distribution parameter $\phi$ is calculated as
\begin{align}
    \nabla_\phi \Lc_{IKL} = \int_{t=0}^T w(t) \mathbb{E}_{q_{\phi}(\xbf_t|\ybf)} \left[ \nabla_{\xbf_t} \log q_{\phi}(\xbf_t|\ybf) - \nabla_{\xbf_t} \log p(\xbf_t) \right] \frac{d \xbf_t}{d\phi} dt. \label{eq:22}
\end{align}

\subsection{Analytical derivation of the gradient of IKL divergence}
Here, we outline the analytical derivation of Eq.~\ref{eq:22}, focusing on an alternative posterior distribution $q_\phi(\xbf_0|\ybf)$, rather than leveraging a neural network for parameterization as in our proposed method.

\noindent \textbf{Gaussian distribution.}
We explore the scenario where the posterior distribution $q_\phi(\xbf_0 | \ybf)$ is modeled as a Gaussian distribution, characterized by a learnable mean $\mu_\phi$ and a fixed variance $\sigma$, \ie, $q_\phi(\xbf_0 | \ybf) = \mathcal{N}(\xbf_0; \ \mu_\phi, \sigma^2\mathbf{I})$. Given the Gaussian nature of $q_\phi(\xbf_0|\ybf)$ and the forward diffusion process $q(\xbf_t|\xbf_0)$ with a Gaussian transition kernel defined as $\xbf_t = \sqrt{\alpha_t}\xbf_0 + \sqrt{(1-\alpha_t)}\epsilon$, $\epsilon \sim \mathcal{N}(\mathbf{0}, \mathbf{I})$, it follows that $q_\phi(\xbf_t|\ybf)$ is also a Gaussian. Consequently, $q_\phi(\xbf_t|\ybf)$ can be expressed as $q_\phi(\xbf_t|\ybf) = \mathcal{N}(\xbf_t; \ \sqrt{\alpha_t}\mu_\phi, (\alpha_t\sigma^2 + (1-\alpha_t))\mathbf{I})$, since $q_\phi(\xbf_t|\ybf) = \int q(\xbf_t|\xbf_0)q_\phi(\xbf_0|\ybf) d\xbf_0$. 
Then, $\nabla_{\xbf_t} \log q_\phi(\xbf_t|\ybf)$ can be analytically calculated as 
\begin{align}
    \nabla_{\xbf_t} \log q_\phi(\xbf_t|\ybf) = -\frac{\epsilon}{\sqrt{\alpha_t\sigma^2 + (1-\alpha_t)}}. \label{eq:23}
\end{align}
Given Eq.~\eqref{eq:22} and \eqref{eq:23}, the gradient of $\Lc_{IKL}$ with respect to $\phi$ is calculated as
\begin{align}
    \nabla_{\phi} & \Lc_{IKL} \\ & = \int_{t=0}^T w(t) \mathbb{E}_{q_{\phi}(\xbf_t|\ybf)} \left[ \nabla_{\xbf_t} \log q_{\phi}(\xbf_t|\ybf) - \nabla_{\xbf_t} \log p(\xbf_t) \right] \frac{d \xbf_t}{d\phi} dt \\
    & \approx \int_{t=0}^T w(t) \mathbb{E}_{\epsilon, \xbf_0} \left[-\frac{\epsilon}{\sqrt{\alpha_t\sigma^2 + (1-\alpha_t)}} - s_\theta(\xbf_t(\xbf_0, \epsilon, t), t) \right] \frac{d \xbf_t(\xbf_0, \epsilon, t)}{d\phi} dt, \label{eq:26}
\end{align}
where $\xbf_t(\xbf_0, \epsilon, t) = \sqrt{\alpha_t}\mu_\phi + \sqrt{\alpha_t \sigma^2 + (1-\alpha_t)} \epsilon$ and $\nabla_{\xbf_t} \log p(\xbf_t)\approx s_\theta(\xbf_t, t)$. By reparameterizing the pre-trained diffusion model $s_\theta$ in a noise prediction form, \ie, $s_\theta(\xbf_t(\xbf_0, \epsilon, t), t) = -\frac{\epsilon_\theta(\xbf_t(\xbf_0, \epsilon, t), t)}{\sqrt{1-\alpha_t}}$, Eq.~\eqref{eq:26} can be rewritten as
\begin{align}
    \nabla_{\phi} & \Lc_{IKL} \\ & = \int_{t=0}^T w(t) \mathbb{E}_{\epsilon, \xbf_0} \left[- \frac{\epsilon}{\sqrt{\alpha_t\sigma^2 + (1-\alpha_t)}} + \frac{\epsilon_\theta(\xbf_t(\xbf_0, \epsilon, t), t)}{\sqrt{1-\alpha_t}} \right] \frac{d \xbf_t(\xbf_0, \epsilon, t)}{d\phi} dt \\ 
    & = \int_{t=0}^T w'(t) \mathbb{E}_{\epsilon, \xbf_0} \left[\eta_t \epsilon - \epsilon_\theta(\xbf_t(\xbf_0, \epsilon, t), t) \right] \frac{d \xbf_t(\xbf_0, \epsilon, t)}{d\phi} dt \label{eq:27},
\end{align}
where $w'(t) = -\frac{w(t)}{\sqrt{1-\alpha_t}}$ and $\eta_t = \frac{\sqrt{1-\alpha_t}}{\sqrt{\alpha_t\sigma^2 + (1-\alpha_t)}}$.
\\

\noindent \textbf{Dirac distribution.}
If we suppose $\sigma = 0$, \ie, $q_\phi(\xbf_0|\ybf) = \delta(\xbf_0 - \mu_\phi)$, Eq.~\eqref{eq:26} can be rewritten as 
\begin{align}
    \nabla_\phi \Lc_{IKL} & = \int_{t=0}^T \frac{w(t)}{\sqrt{1-\alpha_t}} \mathbb{E}_{\epsilon, \xbf_0} \left[- \epsilon + \epsilon_\theta(\xbf_t(\xbf_0, \epsilon, t), t) \right] \frac{d \xbf_t(\xbf_0, \epsilon, t)}{d\phi} dt \\ 
    & = \int_{t=0}^T w'(t) \mathbb{E}_{\epsilon, \xbf_0} \left[\epsilon - \epsilon_\theta(\xbf_t(\xbf_0, \epsilon, t), t) \right] \frac{d \xbf_t(\xbf_0, \epsilon, t)}{d\phi} dt \label{eq:29},
\end{align}
where $w'(t) = -\frac{w(t)}{\sqrt{1-\alpha_t}}$. Note that Eq.~\eqref{eq:29} corresponds to Score Distillation Sampling (SDS) loss~\cite{poole2022dreamfusion}, which is utilized in RED-diff~\cite{mardani2023variational}.
\\

\noindent \textbf{Comparison with RED-diff.}
RED-diff is also based on variational inference and can be viewed as a special case in our framework by assuming the posterior is a Dirac distribution.
Minimizing the variational objective with a single point tends to seek the mode of the true posterior~\cite{wang2023prolificdreamer}.
This may lead to suboptimal sample quality, as it cannot fully capture the complex functions inherent in inverse problems.
In contrast, our method optimizes diverse samples generated from an implicit posterior distribution parameterized by a neural network. 
This allows DAVI to capture the complex posterior distribution and explore plausible solutions flexibly, as experimentally demonstrated by the superior performance of DAVI against RED-Diff.

%% file: Section_supple/B_additional_anlaysis.tex
\input{Tables_supple/supple_deno}
\input{Tables_supple/supple_color}

\section{Additional Analysis}
\label{sec:supple_addtional}

\subsection{Additional image restoration tasks}
\label{sec:supple_deno_color}
\noindent \textbf{Denoising and Colorization.}
To demonstrate that our proposed method can be applied to a wide range of inverse problems, we extend the evaluation to two additional image restoration tasks: denoising and colorization.
For the implementation details, refer to Section~\ref{sec:implementation}. 
Tab.~\ref{tab:supple_deno} and Tab.~\ref{tab:suppe_color} demonstrate the superior performance of DAVI against baselines.
Specifically, we outperform all baselines across all metrics on the denoising task.
We also showcase superior performance in colorization.
For the denoising task, DAVI achieves a remarkable improvement in the FID score, surpassing the next best method DPS by 20.09 points on the FFHQ and DDRM by 27.09 points on the ImageNet.
For the colorization task, we demonstrate substantial performance gains in the FID metric, notably outperforming the second-best method DDRM by 15.75 points on the FFHQ and 6.13 points on the ImageNet, respectively.
Additionally, we provide qualitative comparisons on the FFHQ in Fig.~\ref{fig_supple:quality_ffhq_deno} and Fig.~\ref{fig_supple:quality_ffhq_color}, and ImageNet dataset in Fig.~\ref{fig_supple:quality_imagenet_deno} and Fig.~\ref{fig_supple:quality_imagenet_color}, respectively.
Both quantitative and qualitative results confirm our efficacy in providing more realistic images.
\\
\input{Tables_supple/supple_fps}
\input{Tables_supple/supple_poisson}

\noindent \textbf{Comparison with FPS.} 
FPS~\cite{dou2024diffusion} elegantly bridges Bayesian posterior sampling and Bayesian filtering using diffusion models.
FPS represents a distribution using multiple samples.
Hence, as the number of particles increases, its performance improves but entails additional computational cost, resulting in slower inference than other baselines.
In contrast, our method implicitly represents the posterior distribution by a neural network and learns it via amortization.
Therefore, at test time, our method requires only one forward pass of the neural network, enabling faster inference compared to FPS. 
We provide additional experimental results to empirically compare the proposed method with FPS.
Tab.~\ref{tab:supple_deno} and Tab.~\ref{tab:supple_fps} show that DAVI outperforms, baselines including FPS in noisy inverse problems.
We provide the qualitative results in Fig.~\ref{fig_supple:quality_ffhq_deno} and Fig.~\ref{fig_supple:quality_imagenet_deno}.
\\
\noindent \textbf{Poisson noise.} 
We also evaluate our framework in the case where the measurements are  contaminated with Poisson noise, a common real-world scenario.
While the overall framework is maintained, we introduce a modification in the data consistency loss to accommodate the characteristics of Poisson noise, following the approach outlined in \cite{chung2023diffusion}:
\begin{equation}
    \Lc_{C} \approx \gamma {\lVert \ybf - \Hbf \xbf_0\lVert}^2_\mathbf{\Lambda}, \ \ [ \mathbf{\Lambda} ]_{ii} \triangleq 1/2 \ybf_j,
   %\label{eq:loss_consistency}
\end{equation}
where $j$ indicates the measurement bin and $\lVert \textbf{a} \lVert^2_\mathbf{\Lambda} = \textbf{a}^T \mathbf{\Lambda} \textbf{a}$. The results, as presented in Tab.~\ref{tab:supple_poisson}, validate the robustness of our proposed framework, affirming its effectiveness across a spectrum of measurement noises, including the inherently challenging Poisson noise.

\input{Tables_supple/supple_generated_data}
\subsection{Training with generated data from diffusion prior}
\label{sec:supple_daviG}
In scenarios where privacy concerns or data availability restrict the collection of sufficient real data, it may be difficult to fully exploit the advantages of our framework. 
To address this challenge, we propose to use generated data from a pre-trained diffusion model as a novel solution.
By sampling from the pre-trained diffusion model, we can generate fake data that resemble the features and distribution of real data, thus preserving privacy while ensuring data diversity and quality. This generated data serves as training data for our implicit posterior distribution, which we refer to as DAVI-G. As outlined in Tab.~\ref{tab:suppe_generated_data}, the results show that DAVI-G maintains competitive performance against the baseline methods. This approach not only addresses privacy and data scarcity concerns but also underscores the versatility and adaptability of our framework in leveraging generated data for effective training.

\subsection{Zero-shot capability}
\label{sec:supple_zeroshot}
\noindent \textbf{Out-of-distribution.}
To explore an out-of-distribution (OOD) data scenario, we first train our framework on source datasets (FFHQ and ImageNet) and apply it to unseen datasets (CelebA-HQ and GoPro) in a zero-shot manner.
Specifically, DAVI, initially optimized on the FFHQ dataset for Gaussian deblurring task, is applied to images from the CelebA-HQ dataset~\cite{karras2017progressive}.
Similarly, a version of DAVI optimized on the ImageNet dataset to solve Gaussian deblurring task is tested on the GoPro dataset~\cite{nah2017deep}. 
We employ the same Gaussian blur operator used for the Gaussian deblurring task to make degraded images with additive Gaussian noise. 
The qualitative results in Fig.~\ref{fig_supple:quality_celeba} and Fig.~\ref{fig_supple:quality_gopro} clearly show that our method addresses inverse problems on unseen measurements without further optimization.
Despite the scene-level images in the GoPro dataset, we provide more realistic solutions across various content types.
Note that this is achieved without requiring additional optimization for each measurement, underscoring DAVI's significant generalization capability on OOD data.

\input{Figures_supple/blind_inverse_problem}

\noindent \textbf{Unknown degradation operator.}
Although blind inverse problems with unknown degradation models are beyond the scope of our work, we evaluate our method on the box inpainting task with unseen masks as illustrated in Fig.~\ref{fig:supple_blind_box}.
During amortized training, we employ box masks sized $128\times128$ placed randomly and inject the additive Gaussian noise. At test time, we restore measurements degraded with a centered $196\times196$ mask with additive Gaussian noise.
Our generalization capabilities enable the restoration even when the operator is unknown.
Note that the baseline methods in this paper require degradation operators and utilize them explicitly in the optimization. Hence, none of the baseline methods specifically designed for non-blind inverse problems are even applicable.

\input{Tables_supple/supple_cddb}

\subsection{Comparison with Image-to-Image translation methods}
\label{sec:supple_i2i}
Here, we compare our proposed method with the current image-to-image translation models, namely I$^2$SB~\cite{liu20232} and CDDB~\cite{chung2024direct}, to broaden the scope of our comparative analysis in Tab.~\ref{tab:supple_cddb}. These models explore a non-linear diffusion model which defines optimal transport between two arbitrary distributions.
In particular, I$^2$SB introduces a novel mathematical framework to efficiently learn non-linear diffusions and employs DDPM sampling to solve the inverse problem. CDDB, building on the framework of I$^2$SB, refines the sampling algorithm to make the model more consistent with the given measurement. Despite their innovative approaches, both models require multiple NFEs, from 20 to 100, to produce visually appealing results, whereas our proposed method, DAVI, only requires a single network evaluation. 

We compare our method against I$^2$SB and CDDB on 4$\times$ super-resolution task in Tab.~\ref{tab:supple_cddb} and Fig.~\ref{fig_supple:quality_cddb}, with implementation details of I$^2$SB and CDDB in Section~\ref{sec:supple_baseline}.
As shown in Tab.~\ref{tab:supple_cddb}, DAVI outperforms image-to-image translation models in all metrics.
While these models achieve comparable FID scores to DAVI using at 20 NFEs, our proposed method showcases outstanding performance, evidenced by 9.6 points FID score improvement over CDDB using 20 NFEs on the ImageNet dataset.
Moreover, at a single NFE, they experience a significant drop in restored image quality, illustrated by I$^2$SB's FID score degradation by 37.58 points compared to 100 NFEs on the FFHQ dataset.
In contrast, DAVI maintains remarkable quality with just a single network evaluation, highlighting the efficacy of our framework.

\subsection{Limitations}
Our method exhibits impressive zero-shot generalization capabilities utilizing amortized variational inference, yet it's important to acknowledge that the extent of this generalization is affected by both the capacity and expressiveness of the chosen neural network. 
However, the amortization-based method requires training unlike optimization-based baselines.
Additionally, the inherent nature of amortized variational inference, which optimizes a shared approximation across different inputs, might not always perfectly represent the unique attributes of each data point.
Thus, a more detailed examination of the variational family selection and neural network architecture could potentially amplify our method's generalization ability. We believe that exploring these avenues remains a promising direction for future research.

%% file: Tables_supple/supple_deno.tex
\begin{table*}[t]
\centering
\caption{\textbf{Results on denoising task}. The highest performance is highlighted in bold, while the second-highest one is underlined.}
\resizebox{0.95\linewidth}{!}{
\begin{tabular}{l|c|ccccccc} 
\hline
       &             & \multicolumn{3}{c}{FFHQ 256$\times$256}      &  & \multicolumn{3}{c}{ImageNet 256$\times$256}  \\
Method      & NFE $\downarrow$   & PSNR $\uparrow$ & LPIPS $\downarrow$ & FID $\downarrow$ &  & PSNR $\uparrow$ & LPIPS $\downarrow$ & FID $\downarrow$ \\ 
\hline
\hline
DDRM~\cite{kawar2022denoising} & 20 &           30.35 & 0.222 & 58.95 & & 22.96 & 0.361 & \underline{40.47} \\
DDNM$^+$\cite{wang2022zero} & 100 &       \underline{31.68} 	& \underline{0.210}	& 63.49 & & \underline{29.46} & \underline{0.262} & 63.54 \\
DPS~\cite{chung2023diffusion} & 1000 &           23.68 & 0.299 & \underline{42.94} & & 21.83 & 0.449 & 76.01 \\
$\Pi$GDM~\cite{song2023pseudoinverse} & 100 &       22.30 & 0.437 & 62.89 & & 22.22 & 0.401 & 51.61 \\
RED-diff~\cite{mardani2023variational} & 1000 &        20.44 & 0.465 & 64.74& & 20.45 & 0.430 & 52.43 \\
FPS~\cite{dou2024diffusion} & 1000 &         26.04 & 0.304 & 44.45 & & 20.85 & 0.426 & 98.84 \\
\hline
\textbf{DAVI (Ours)} & 1 &         \textbf{31.72} & \textbf{0.131} & \textbf{22.85} & & \textbf{31.57} & \textbf{0.119} & \textbf{13.38} \\
\hline
\end{tabular}
}
\label{tab:supple_deno}
\end{table*}

%% file: Tables_supple/supple_color.tex
\begin{table*}[t]
\centering
\caption{\textbf{Results on colorization task}. The highest performance is highlighted in bold, while the second-highest one is underlined.}
\resizebox{0.95\linewidth}{!}{
\begin{tabular}{l|c|ccccccc} 
\hline
       &             & \multicolumn{3}{c}{FFHQ 256$\times$256}      &  & \multicolumn{3}{c}{ImageNet 256$\times$256}  \\
Method      & NFE $\downarrow$   & PSNR $\uparrow$ & LPIPS $\downarrow$ & FID $\downarrow$ &  & PSNR $\uparrow$ & LPIPS $\downarrow$ & FID $\downarrow$ \\ 
\hline
\hline
DDRM~\cite{kawar2022denoising}  & 20 &           22.87 & \underline{0.209} & \underline{43.82} &   & 22.26 & \underline{0.243} & \underline{47.17}  \\
DDNM$^+$\cite{wang2022zero}   & 100 &            \underline{23.85} & 0.230 & 58.53 &             & 21.38 & 0.310 & 78.13 \\
DPS~\cite{chung2023diffusion} & 1000 &           17.27 & 0.320 & 60.15 &             & 17.67 & 0.578 & 93.25 \\
$\Pi$GDM~\cite{song2023pseudoinverse}   & 100 &           21.66	& 0.254	& 45.47 &    & 20.58 & 0.276 & 54.01 \\
RED-diff~\cite{mardani2023variational} & 1000 &           23.61 & 0.331 & 63.38 &    & \textbf{22.51} & 0.353 & 66.72 \\
\hline
\textbf{DAVI (Ours)}  & 1 &        \textbf{25.35}    &  \textbf{0.169}  & \textbf{28.07} &     &  \underline{22.27}  &  \textbf{0.202}  &  \textbf{41.04} \\
\hline
\end{tabular}
}
\label{tab:suppe_color}
\end{table*}

%% file: Tables_supple/supple_fps.tex
\begin{table}[!ht]
\centering
\caption{\textbf{Results on quantitative comparison with FPS}. The highest performance is highlighted in bold.}
\resizebox{0.96\linewidth}{!}{
\begin{tabular}{l|ccccccc} 
\hline
                    & \multicolumn{3}{c}{Gaussian deblur}                     &  & \multicolumn{3}{c}{SR($\times$4)}               \\
Method              & PSNR $\uparrow$ & LPIPS $\downarrow$ & FID $\downarrow$ &  & PSNR $\uparrow$ & LPIPS $\downarrow$ & FID $\downarrow$  \\ 
\hline
FPS~\cite{dou2024diffusion}                   & 25.17 & 0.237 & 33.57 & & 26.92 & 0.182 & 25.08 \\
\textbf{DAVI (Ours)}   & \textbf{25.46} & \textbf{0.225} & \textbf{29.92} & & \textbf{28.23} & \textbf{0.171} & \textbf{23.96} \\
\hline
\end{tabular}
}
\label{tab:supple_fps}
\end{table}

%% file: Tables_supple/supple_poisson.tex
\begin{table*}[t]
\centering
\caption{\textbf{Results on poisson noise measurements} (Gaussian deblur and 4$\times$ super-resolution on the FFHQ 256$\times$256). The highest performance is highlighted in bold, while the second-highest one is underlined.}
\resizebox{0.95\linewidth}{!}{%
\begin{tabular}{l|c|cccccccc} 
\hline
       &             & \multicolumn{3}{c}{Gaussian deblur}                     &  & \multicolumn{3}{c}{4$\times$ Super-resolution}   \\
Method      & NFE $\downarrow$   & PSNR $\uparrow$ & LPIPS $\downarrow$ & FID $\downarrow$ &  & PSNR $\uparrow$ & LPIPS $\downarrow$ & FID $\downarrow$                      \\ 
\hline
DDNM$^+$\cite{wang2022zero}         & 100   &   23.57 & 0.369 & 101.33 &       & \underline{27.01}	& 0.283	& 70.75  \\ 
DPS~\cite{chung2023diffusion}       & 1000   &  23.41 &	\underline{0.276} & \underline{32.35} &       & 25.46 & \underline{0.254} & \underline{35.11}   \\ 
$\Pi$GDM~\cite{song2023pseudoinverse}       & 100   &   23.78 & 0.291 & 43.76 &       & 25.70 & 0.276 & 58.44   \\ 
RED-diff~\cite{mardani2023variational}      & 1000   &  \textbf{25.58} & 0.328 & 49.37 &       & 23.39 & 0.497 & 124.34  \\ 
\hline
\textbf{DAVI (Ours)}  & 1 &   \underline{25.22} & \textbf{0.240} & \textbf{32.12} &                   &                    \textbf{27.86}  & \textbf{0.189} &  \textbf{25.72} &        \\
\hline
\end{tabular}
}
\label{tab:supple_poisson}
\end{table*}

%% file: Tables_supple/supple_generated_data.tex
\begin{table*}[t]
\centering
\caption{\textbf{Results with generated data from diffusion model} (4$\times$ super-resolution and box inpainting tasks on FFHQ 256$\times$256). The highest performance is highlighted in bold, while the second-highest one is underlined.}
\resizebox{0.95\linewidth}{!}{%
\begin{tabular}{l|c|ccccccc} 
\hline
       &             & \multicolumn{3}{c}{4$\times$ super-resolution}      &  & \multicolumn{3}{c}{Box inpainting}  \\
Method      & NFE $\downarrow$   & PSNR $\uparrow$ & LPIPS $\downarrow$ & FID $\downarrow$ &  & PSNR $\uparrow$ & LPIPS $\downarrow$ & FID $\downarrow$ \\ 
\hline
\hline
DDRM~\cite{kawar2022denoising}  & 20 &           28.09 & 0.228 & 48.19 &   & 22.27 & \underline{0.177} & 37.62  \\
DDNM$^+$\cite{wang2022zero}   & 100 &            \underline{28.17} & 0.244 & 59.09 &             & 23.79 & 0.218 & 48.82 \\
DPS~\cite{chung2023diffusion} & 1000 &           25.55 & 0.255 & 36.29 &             & 23.79 & 0.218 & 48.82 \\
$\Pi$GDM~\cite{song2023pseudoinverse}   & 100 &           27.73	& \underline{0.225}	& 49.86 &    & 23.04 & 0.220 & 32.99 \\
DiffPIR~\cite{zhu2023denoising}  &    100    & 25.32          & 0.341          & \underline{40.35}                      &  & 23.97 & 0.195 & \underline{31.27}      \\
RED-diff~\cite{mardani2023variational} & 1000 &           26.75 & 0.379 & 92.82 &    & \underline{24.16} & 0.216 & 35.80 \\
\hline
\textbf{DAVI-G (Ours)}  & 1 &          \textbf{28.58}  &  \textbf{0.196} &  \textbf{30.61} &    & \textbf{25.06} & \textbf{0.134} & \textbf{26.66} \\
\hline
\end{tabular}
}
\label{tab:suppe_generated_data}
\end{table*}

%% file: Figures_supple/blind_inverse_problem.tex
\begin{figure}[ht]
    \centering
    \includegraphics[width=1\columnwidth]{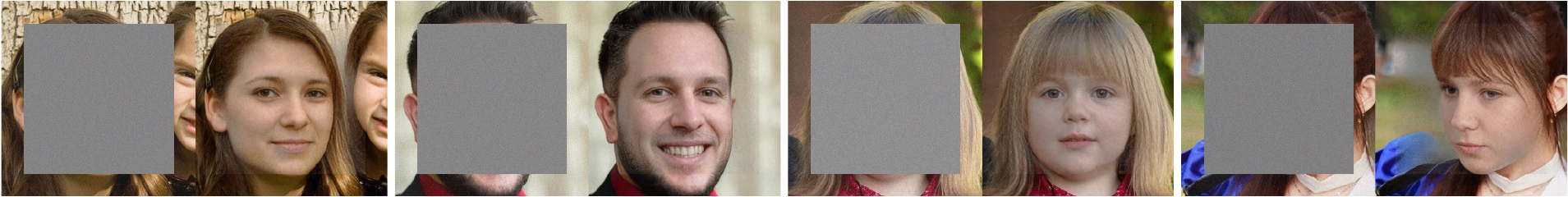}
    \caption{Zero-shot capabilities of unseen mask size $196\times196$ on FFHQ 256$\times$256.}
    \label{fig:supple_blind_box}
\end{figure}

%% file: Tables_supple/supple_cddb.tex
\begin{table*}[t]
\centering
\caption{\textbf{Comparison with image-to-image translation model on 4$\times$ super-resolution}. The highest performance is highlighted in bold, while the second-highest one is underlined.}
\resizebox{0.95\linewidth}{!}{
\begin{tabular}{l|c|ccccccc} 
\hline
       &             & \multicolumn{3}{c}{FFHQ 256$\times$256}                     &  & \multicolumn{3}{c}{ImageNet 256$\times$256}  \\
Method      & NFE $\downarrow$   & PSNR $\uparrow$ & LPIPS $\downarrow$ & FID $\downarrow$ &  & PSNR $\uparrow$ & LPIPS $\downarrow$ & FID $\downarrow$  \\ 
\hline
\hline
                         & 1&  \underline{27.95} & 0.292 & 66.91 &     & \underline{25.99} & \underline{0.330}	& 56.93 \\
I$^2$SB~\cite{liu20232}  & 20 &   27.09 & 0.223 & 28.46 &   & 23.90 & 0.350 & 47.91 \\
                         & 100 &  26.40 & 0.234 & 29.33 &                    & 22.56 & 0.394 & 52.13 \\
\hline
                            & 1&        27.89 & 0.294 & 67.15 &   & 25.97 & 0.331 & 56.80 \\
CDDB~\cite{chung2024direct} & 20 &      27.39 & \underline{0.217} & \underline{27.84} &   & 24.26 & 0.340 & \underline{45.87} \\
                            & 100 &      26.80 & 0.228 & 30.92 &  & 22.85 & 0.384 & 50.30  \\
\hline
\textbf{DAVI (Ours)}        & 1 &          \textbf{28.23}          &  \textbf{0.171}       &  \textbf{23.96}    &                     & \textbf{26.58} &  \textbf{0.242} &  \textbf{36.27}  \\
\hline
\end{tabular}
}
\label{tab:supple_cddb}
\end{table*}

%% file: Section_supple/C_experimental_details.tex
\section{Experimental Details}
\label{sec:supple_details}
\subsection{Implementation details}
\label{sec:implementation}
In our framework, we utilize several hyperparameters, including the $\bar{\sigma}_a$, $p(a)$, perturbation scale $h$, IKL time $T$, weight $\gamma$, regularization coefficient, training iteration $K$, and learning rate for the updates of $\phi$ and $\psi$.
For all experiments, we use a monotonically increasing function for perturbation schedule as $\bar{\sigma}_a = 1 - \alpha_a$, where $\alpha_a = \prod_{i=1}^a \beta_i$, and the beta distribution with the shape parameters of 3 and 1 for $p(a)$, \ie, $p(a) = \frac{1}{B(3,1)}a^{2}(1-a)^0$.
During training time, we empirically select hyperparameters via grid search using training data.
For instance, we search for the perturbation scale $h$ within $[0.01, 0.1]$ to adjust the variance of the samples drawn from PPB.
For more details on hyperparameters, we provide a comprehensive list of those used in each experiment in Tab.~\ref{tab:supple_hyperparameters}.
All the models, including $\Ic_\phi, s_\phi$, and $s_\theta$, are initialized from the ADM~\cite{dhariwal2021diffusion} 256$\times$256 model for faster convergence.
We choose the AdamW optimizer by setting the learning rate to 1e-4 for each parameter $\phi$ and $\psi$, respectively.
We conduct experiments with a batch size of 8 on the FFHQ dataset and 10 or 12 on the ImageNet dataset, using the number of iterations specified in Tab.~\ref{tab:supple_hyperparameters}.
Since the pre-trained diffusion model is trained with noise-prediction form~\cite{ho2020denoising}, we re-parameterize $\Ic_{\phi, a}$ with the score network-induced data-prediction transformation as
\begin{align}
    \xbf_0 = \Ic_{\phi, a}(\ybf_a) = \ybf_{a} + \alpha_a \Ic'_{\phi,a}(\ybf_{a}),
\end{align}
where $\Ic'_{\phi, a}$ is the neural network initialized by the pre-trained diffusion model.
We utilize the 49K FFHQ training dataset and 130K subset of the ImageNet training dataset for amortized optimization, which is a distinct set from the validation dataset used for evaluation.
To stabilize the optimization process of the implicit distribution, we add the $l_2$ norm between the training data and the restored image as a regularization term.
\\
\noindent \textbf{Operator setting.} Following the baselines~\cite{kawar2022denoising,wang2022zero}, we adopt the implementation of all degradation operators from DDRM~\cite{kawar2022denoising}.
For a fair comparison, we evaluate each task with the same operator and the same amout of measurement noise.
\\
\noindent \textbf{Evaluation.} We evaluate the FID between the ground truth images and reconstructed images using the \texttt{pytorch-fid} package.
\\
\noindent \textbf{Computational cost.}
For inference time, DAVI and all baselines use the same diffusion model architecture for each dataset including FFHQ and ImageNet.
As shown in Tab.~\ref{tab:supple_flops}, the FLOPs are proportional to the number of function evaluations (NFEs), resulting in $\times 20\sim \times 1000$ speed-up. 
For training time, similar to other amortization-based methods,  we require the training cost.
We conduct experiments on the FFHQ dataset using 4 TITAN RTX GPUs, whereas those on the ImageNet dataset with 4 RTX A6000 GPUs.
Training time for each task varies as we optimize our implicit model until it converges.

\input{Tables_supple/hyperparameters}
\input{Tables_supple/supple_flops}

\subsection{Baselines}
\label{sec:supple_baseline}
Here, we provide the hyperparameters used in the re-implementation of the baselines: DDRM~\cite{kawar2022denoising}, DDNM~\cite{wang2022zero}, DPS~\cite{chung2023diffusion}, $\Pi$GDM~\cite{song2023pseudoinverse}, DiffPIR~\cite{zhu2023denoising}, RED-Diff~\cite{mardani2023variational}, FPS~\cite{dou2024diffusion}, I$^2$SB~\cite{liu20232} and CDDB~\cite{chung2024direct}. 
The same pre-trained diffusion models from \cite{chung2023diffusion} were used across all methods for a fair comparison of two benchmark datasets FFHQ $256\times 256$ and ImageNet $256\times 256$.
For tasks not covered in the original paper, we tuned hyperparameters to achieve optimal performance.
We applied the additive Gaussian noise $\sigma_\ybf=0.05$ to all measurements to define noisy inverse problems.
\\
\noindent\textbf{DDRM.} For all experiments, we used the default settings specified in the original DDRM paper. We set $\eta_B=1.0$ and $\eta=0.85$, with 20 NFE DDIM sampling. 
\\ 
\noindent\textbf{DDNM.} We employed DDNM$^+$, which is tailored for noisy measurements by implementing the time-travel trick. We set $\eta=0.85$ and 100 NFEs. For the time-travel, we tuned $s$ and $r$ for the best performance. \\
\noindent\textbf{DPS.} We used 1,000 NFEs for all experiments and followed the step size $\zeta$ as provided in its paper. \\
\noindent\textbf{$\Pi$GDM.} To follow our noisy inverse problems setting, we tuned $\eta$, maintaining 100 NFEs. $\Pi$GDM's standard setting of $\eta = 1.0$ is adjusted within the range of $[0.5, 1.0, 1.5, 2.0]$ to yield the best performance. \\
\noindent\textbf{DiffPIR.} Our implementation of DiffPIR followed the hyperparameters of guidance scale $\lambda$ and $\zeta$ as its original paper. \\
\noindent\textbf{RED-diff.} Our implementation of RED-diff followed the default configuration in the original paper with 1,000 NFEs.\\ 
\noindent\textbf{FPS.} We follow the official code\footnote{https://github.com/ZehaoDou-official/FPS-SMC-2023} and used 1,000 NFEs. We set a particle size $M=1$ due to computational demands.\\
\noindent\textbf{I$^2$SB} We follow the official code\footnote{https://github.com/NVlabs/I2SB} for training the model on
the FFHQ dataset. For the ImageNet dataset, we utilize the official pre-trained I$^2$SB model.\\
\noindent\textbf{CDDB} We follow the implementation of the offical code\footnote{https://github.com/HJ-harry/CDDB} to generate samples with $1, 20,$ and $100$ NFEs, using the same model as I$^2$SB.

For the experiment conducted in Section 5.2 of the main paper for robustness to unknown noise scale, we use the noise scale of 0.05 for the algorithms of $\Pi$GDM, DDRM, and DDNM$^+$, which were tuned to yield the optimal performance for each model.

%% file: Tables_supple/hyperparameters.tex
\begin{table}[t]
\centering
\caption{\textbf{Hyperparameters of DAVI.} Detailed hyperparameters of implementation for five restoration tasks and two benchmark datasets.}
\resizebox{0.9\columnwidth}{!}{%
\begin{tabular}{llccccccccc} 
\hline
    &  & Gaussian & & 4$\times$ SR & & Inpainting &  & Denoising & & Colorization  \\ 
\hline
\hline
\multicolumn{11}{l}{{\cellcolor[rgb]{0.899,0.899,0.899}}\textbf{FFHQ 256$\times$256}}                                                        \\
perturbation scale $h$    &      & 0.1          & &   0.1              & &      0.1       &  &  0.1        & & 0.1        \\
IKL time $T$     &      &   400      &  &  1000               & &  1000     & &    1000       & &    1000                \\
weight $\gamma$ &     &  \sj{0.5}         & &  \sj{0.1}               & &   \sj{0.5}         & &   0.2    & &   1.0        \\
regularization coeff  &   &  0.25 &              &  1.0 &               &  1.0    & &  0.1    & &  0.25   \\
iteration $K$  &   &  42k &              &  40k &               &  24k    & &  48k    & &  22k   \\
batch size  &   &  8 &              &  8 &               &  8    & &  8    & &  8   \\
\hline

\multicolumn{11}{l}{{\cellcolor[rgb]{0.899,0.899,0.899}}\textbf{ImageNet 256$\times$256}}                                                    \\
perturbation scale $h$   &      & 0.1          & &   0.01              & &      0.01       &  &  0.01   & & 0.01            \\
IKL time $T$     &      &   400     &   &  1000              &  &  400            & &    600  & &    1000                \\
weight $\gamma$ &     &  0.5       &   &  0.075             &  &   0.01           & &   0.1  & &   0.5    \\
regularization coeff  &   &  0.5 &              &  0.25 &               &  0.1   & &  1.0    & & 0.1   \\
iteration $K$     &      &   144k     &   &  6k              &  &  189k           & &    6k  & &    45k                \\
batch size  &   &  12 &              &  10 &               &  10    & &  12    & &  10   \\
\hline
\end{tabular}
}
\label{tab:supple_hyperparameters}
\end{table}

%% file: Tables_supple/supple_flops.tex
\begin{table}[!ht]
\centering
\caption{\textbf{GFLOPs of Inference time.}. We calculate computational cost for 4$\times$ super-resolution task on FFHQ 256$\times$256 dataset. The highest performance is highlighted in bold. FLOPs are proportional to the number of NFE.}
\resizebox{\linewidth}{!}{
\begin{tabular}{l|c|c|lcc|lcc} 
\hline
Method & \textbf{DAVI (Ours)} & DDRM & DDNM$^+$ & $\Pi$GDM & DiffPIR  & DPS & RED-diff & FPS \\
\hline
NFE $\downarrow$ & \textbf{1} & 20 & \multicolumn{3}{c|}{100} & \multicolumn{3}{c}{1000} \\
FLOPs $\downarrow$ & \textbf{194.7\ G} & 3,893.8 G & \multicolumn{3}{c|}{19,468.9 G} & \multicolumn{3}{c}{194,688.5 G} \\
\hline
\end{tabular}
}
\label{tab:supple_flops}
\end{table}

%% file: Section_supple/D_more_relatedworks.tex
\section{More related works.}
\label{sec:supple_related_works}
In this section, we discuss the concurrent score distillation method based on the diffusion prior.
Using particle-based variational inference, ProlificDreamer~\cite{wang2023prolificdreamer} maintains 3D parameters represented as particles to model the 3D distribution.
To optimize the 3D scene distribution, they minimize the KL divergence, ensuring that the rendered image distribution from any view closely matches the distribution defined by a pretrained 2D diffusion model.
DMD~\cite{yin2024one} combines score distillation loss with regression loss to match the distribution of synthetic data to the real distribution from a pretrained diffusion model. By minimizing the distribution matching objective, they move the generated data toward the modes of the clean distribution of the diffusion prior.
To our best knowledge, DAVI is the first work that studies the implicit posterior distribution using amortized variational inference in the literature. 
In addition, we enhanced the generalization ability by introducing Perturbed Posterior Bridge in the diffusion process.
Different domains and our novel components show the distinction between the related works and our proposed method.

%% file: Section_supple/E_qualitative_results.tex
\section{Additional qualitative results}
\label{sec:supple_qual}
In this section, we compare the results of DAVI with the state-of-the-art baselines for noisy inverse problems.
Fig.~\ref{fig_supple:quality_celeba} and Fig.~\ref{fig_supple:quality_gopro} demonstrate our generalization results through amortized variational inference.
Fig.~\ref{fig_supple:quality_cddb} emphasizes the improved results of our framework DAVI, compared to Image-to-Image translation methods.
From Fig.~\ref{fig_supple:quality_ffhq_gauss} to \ref{fig_supple:quality_ffhq_boxinpaint} and Fig.~\ref{fig_supple:quality_imagenet_gauss} to \ref{fig_supple:quality_imagenet_boxinpaint}, we provide additional qualitative results on image restoration tasks conducted in the main paper, including Gaussian deblurring, 4$\times$ super-resolution, and box inpainting.
Fig.~\ref{fig_supple:quality_ffhq_deno} and Fig.~\ref{fig_supple:quality_imagenet_deno} compare qualitative results on the denoising task.
Fig.~\ref{fig_supple:quality_ffhq_color} and Fig.~\ref{fig_supple:quality_imagenet_color} showcase the qualitative results on colorization task.

\input{Figures_supple/quality_celeba}
\input{Figures_supple/quality_gopro}
\input{Figures_supple/quality_cddb}
\input{Figures_supple/quality_ffhq_gauss}
\input{Figures_supple/quality_ffhq_sr}
\input{Figures_supple/quality_ffhq_boxinpaint}
\input{Figures_supple/quality_ffhq_deno}
\input{Figures_supple/quality_ffhq_color}
\input{Figures_supple/quality_imagenet_gauss}
\input{Figures_supple/quality_imagenet_sr}
\input{Figures_supple/quality_imagenet_boxinpaint}
\input{Figures_supple/quality_imagenet_deno}
\input{Figures_supple/quality_imagenet_color}

%% file: Figures_supple/quality_celeba.tex
\begin{figure*}[p]
    \centering
    \includegraphics[width=0.99\textwidth]{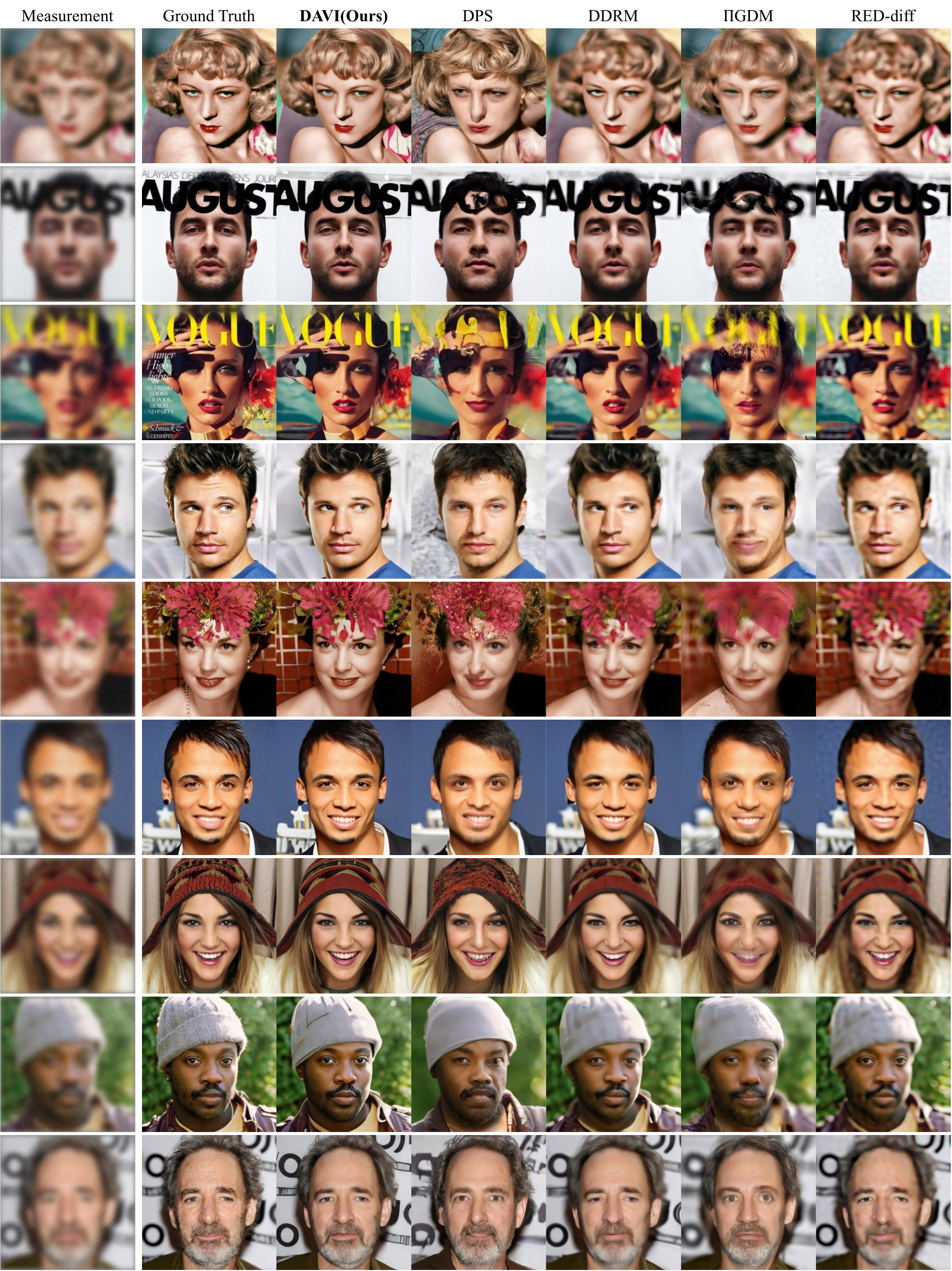}
    \caption{\textbf{Qualitative comparison of Generalization.} Using an FFHQ-trained model, we inference on the CelebA-HQ 256$\times$256 dataset for Gaussian deblurring. DAVI demonstrates superior generalization ability without test-time optimization. DAVI shows the most vivid and realistic solutions while maintaining intricate details of measurement, in contrast to baselines which struggle to satisfy both aspects.}
    \label{fig_supple:quality_celeba}
\end{figure*}

%% file: Figures_supple/quality_gopro.tex
\begin{figure*}[p]
    \centering
    \includegraphics[width=0.99\textwidth]{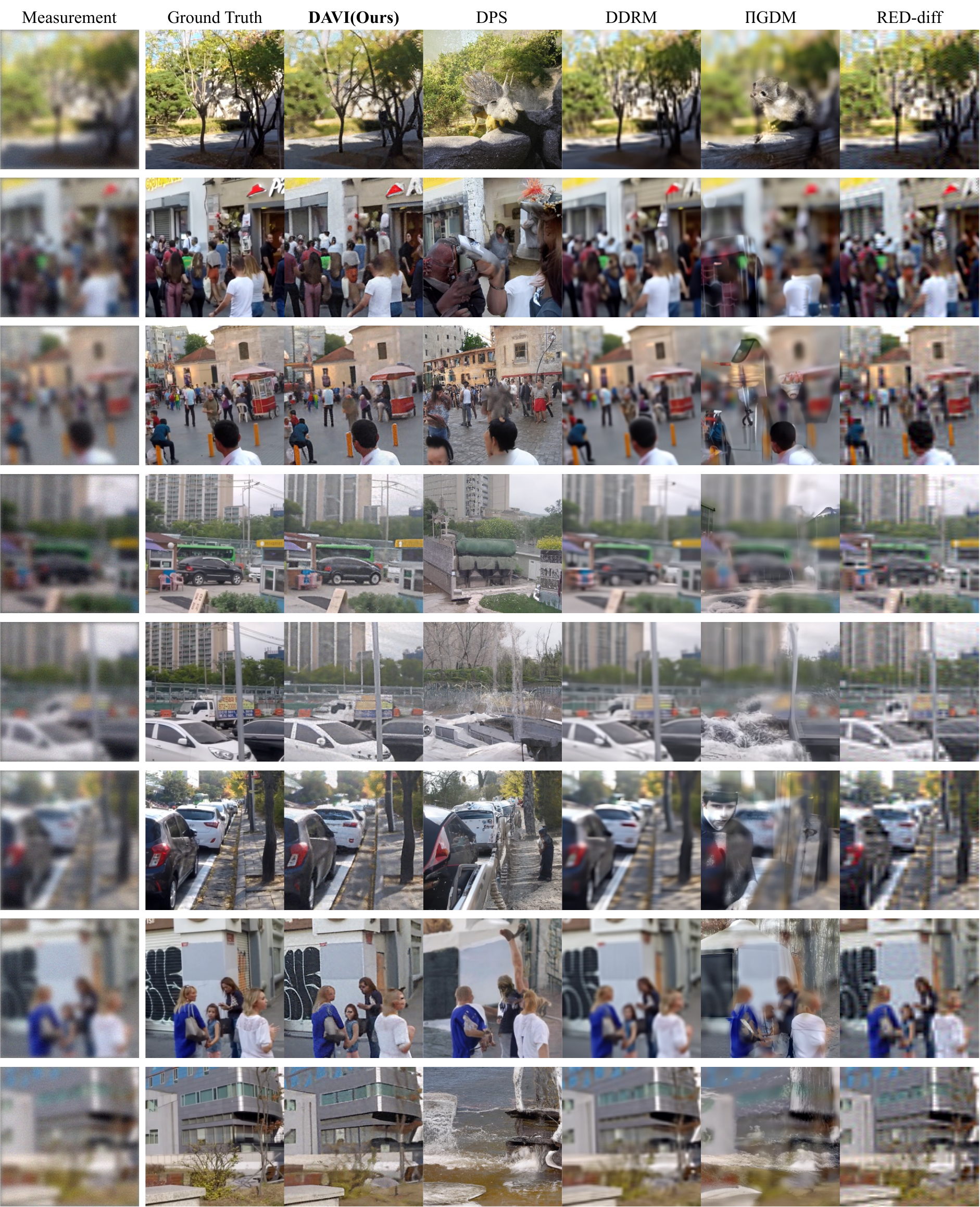}
    \caption{\textbf{Qualitative comparison of Generalization.} Using an ImagneNet-trained model, we inference on the GoPro dataset for Gaussian deblurring. DAVI demonstrates superior generalization ability without test-time optimization. DAVI shows the most realistic solutions while maintaining intricate details of measurement, in contrast to baselines which struggle to satisfy consistency with measurements.}
    \label{fig_supple:quality_gopro}
\end{figure*}

%% file: Figures_supple/quality_cddb.tex
\begin{figure*}[p]
    \centering
    \includegraphics[width=0.87\textwidth]{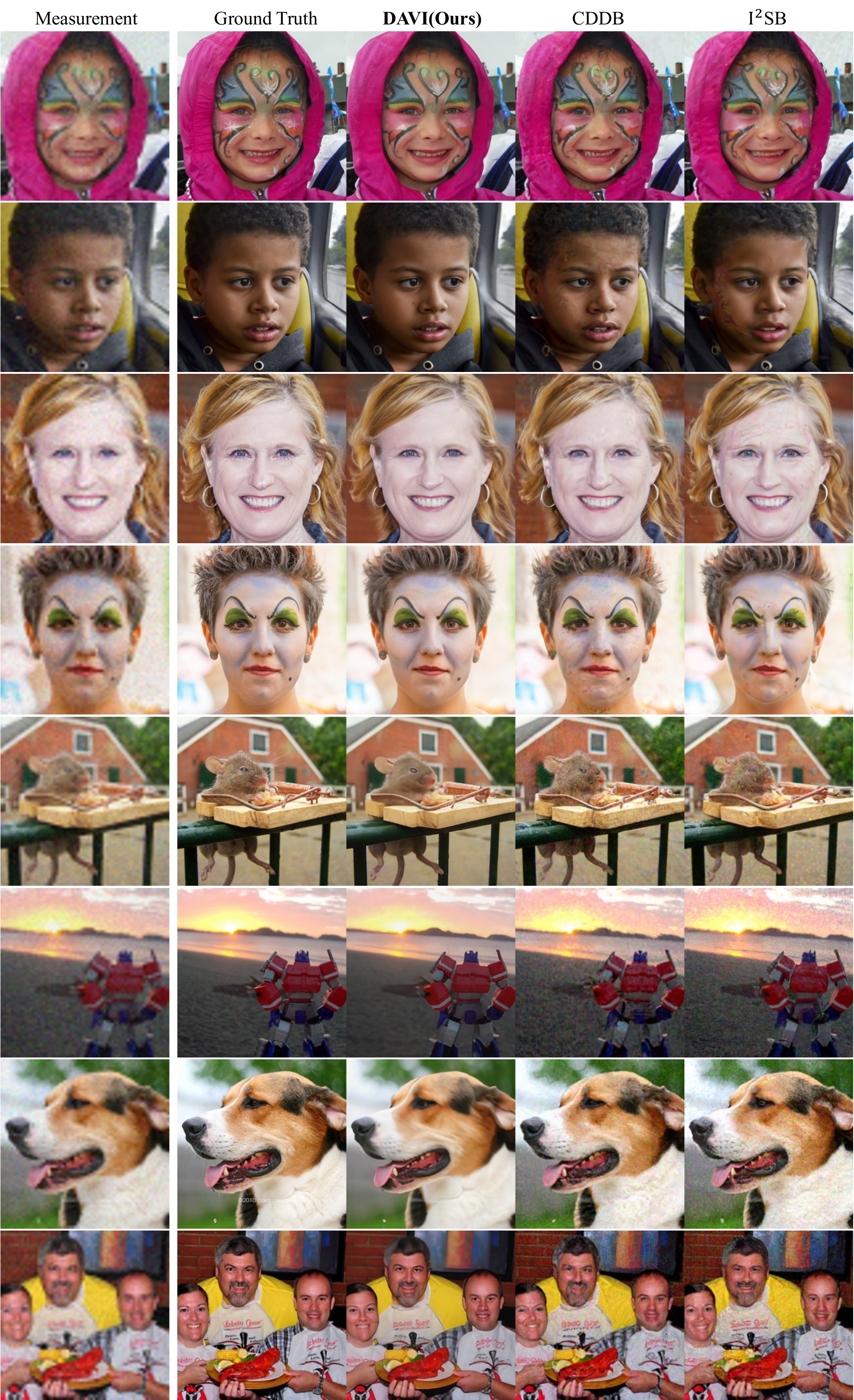}
    \caption{\textbf{Qualitative comparison with Image-to-Image translation methods.} Results for 4$\times$ super-resolution on FFHQ 256$\times$256 (top 4 rows) and ImageNet 256$\times$256 (bottom 4 rows).
    We compare DAVI with 1 NFE to CDDB and I$^2$SB with 20 NFEs.}
    \label{fig_supple:quality_cddb}
\end{figure*}

%% file: Figures_supple/quality_ffhq_gauss.tex
\begin{figure*}[p]
    \centering
    \includegraphics[width=0.99\textwidth]{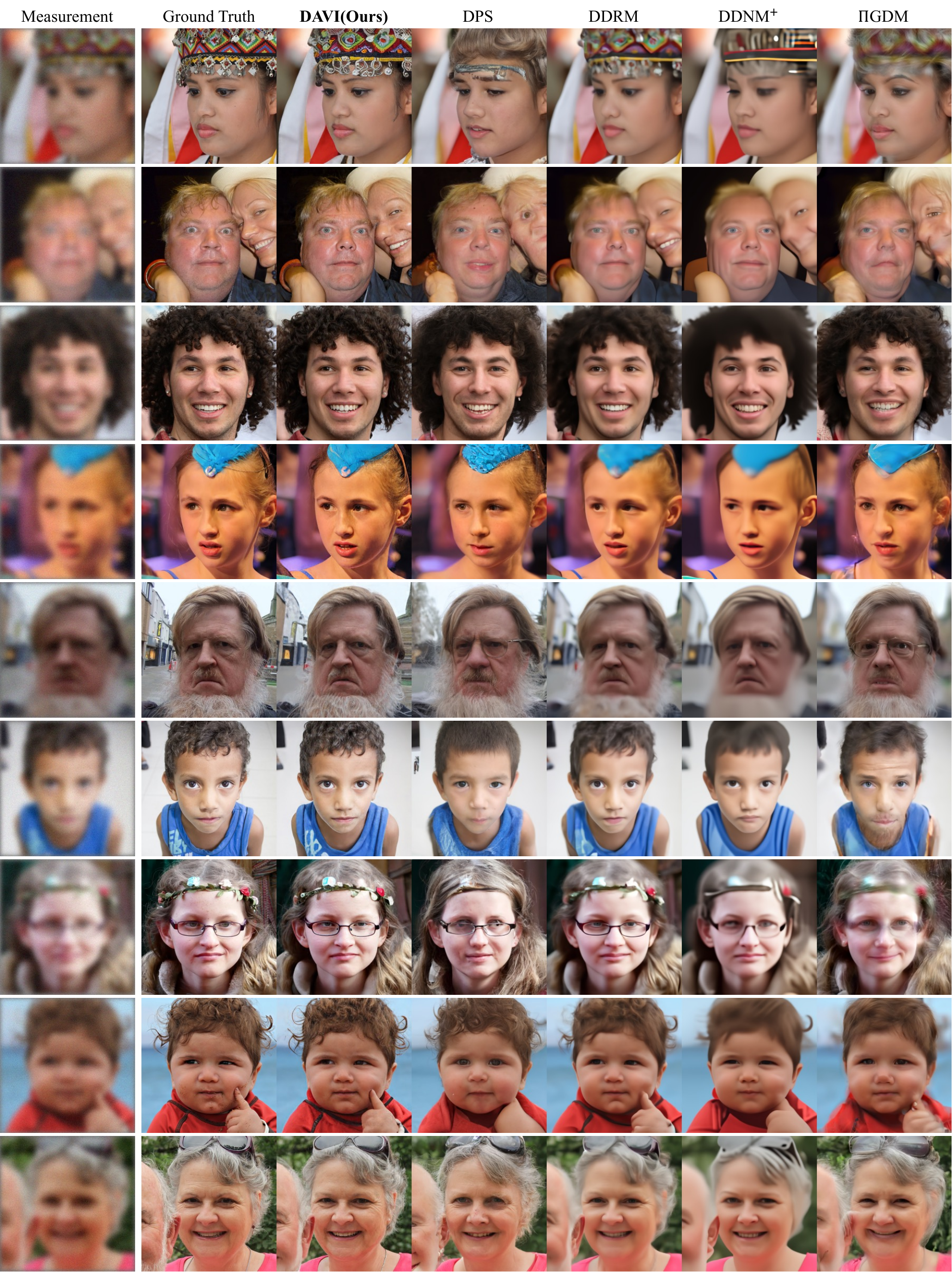}
    \caption{\textbf{Qualitative comparison of noisy inverse problem methods.} Results for Gaussian deblurring on FFHQ 256$\times$256.}
    \label{fig_supple:quality_ffhq_gauss}
\end{figure*}

%% file: Figures_supple/quality_ffhq_sr.tex
\begin{figure*}[p]
    \centering
    \includegraphics[width=0.99\textwidth]{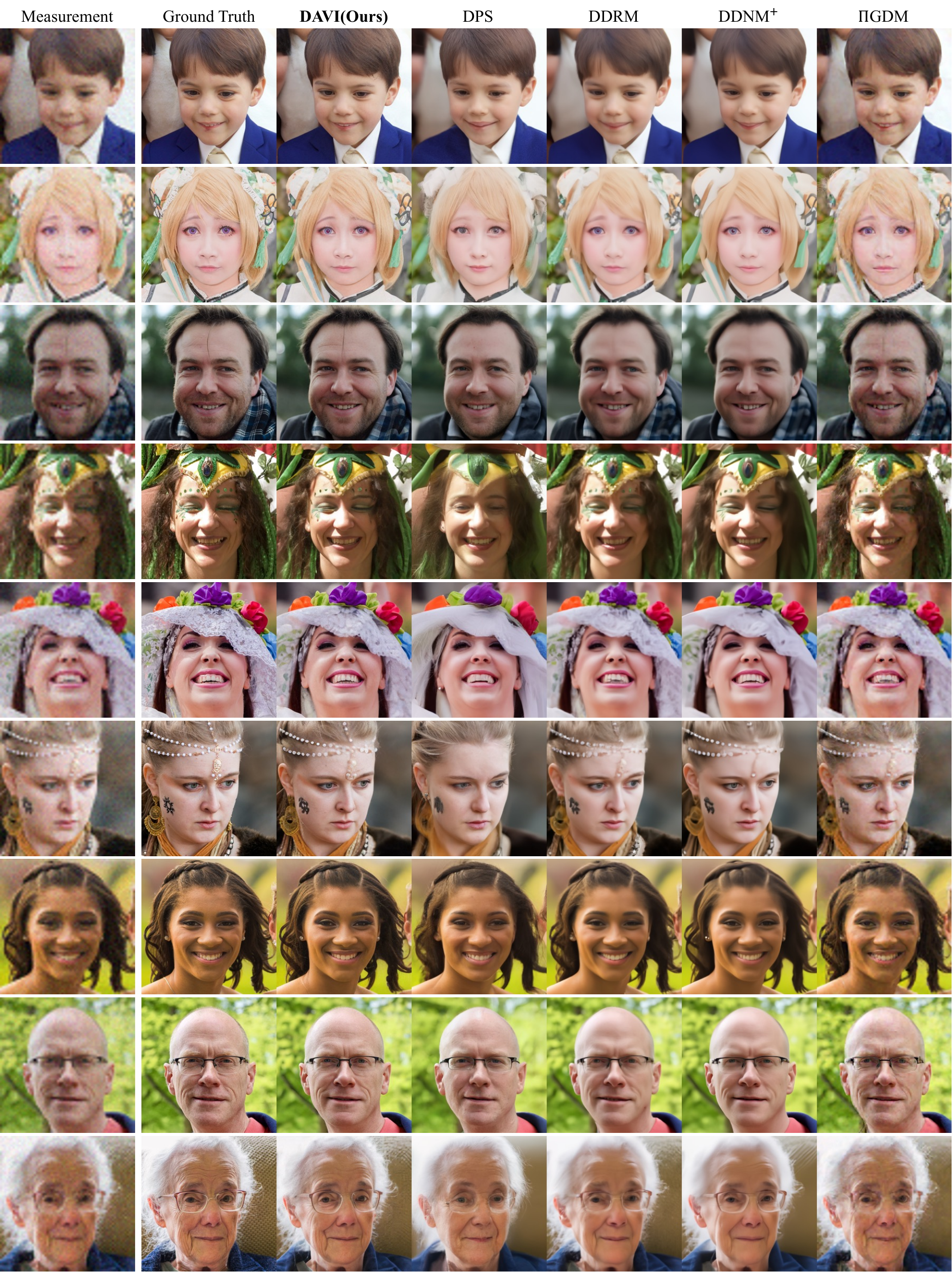}
    \caption{\textbf{Qualitative comparison of noisy inverse problem methods.} Results for 4$\times$ super-resolution on FFHQ 256$\times$256.}
    \label{fig_supple:quality_ffhq_sr}
\end{figure*}

%% file: Figures_supple/quality_ffhq_boxinpaint.tex
\begin{figure*}[p]
    \centering
    \includegraphics[width=0.99\textwidth]{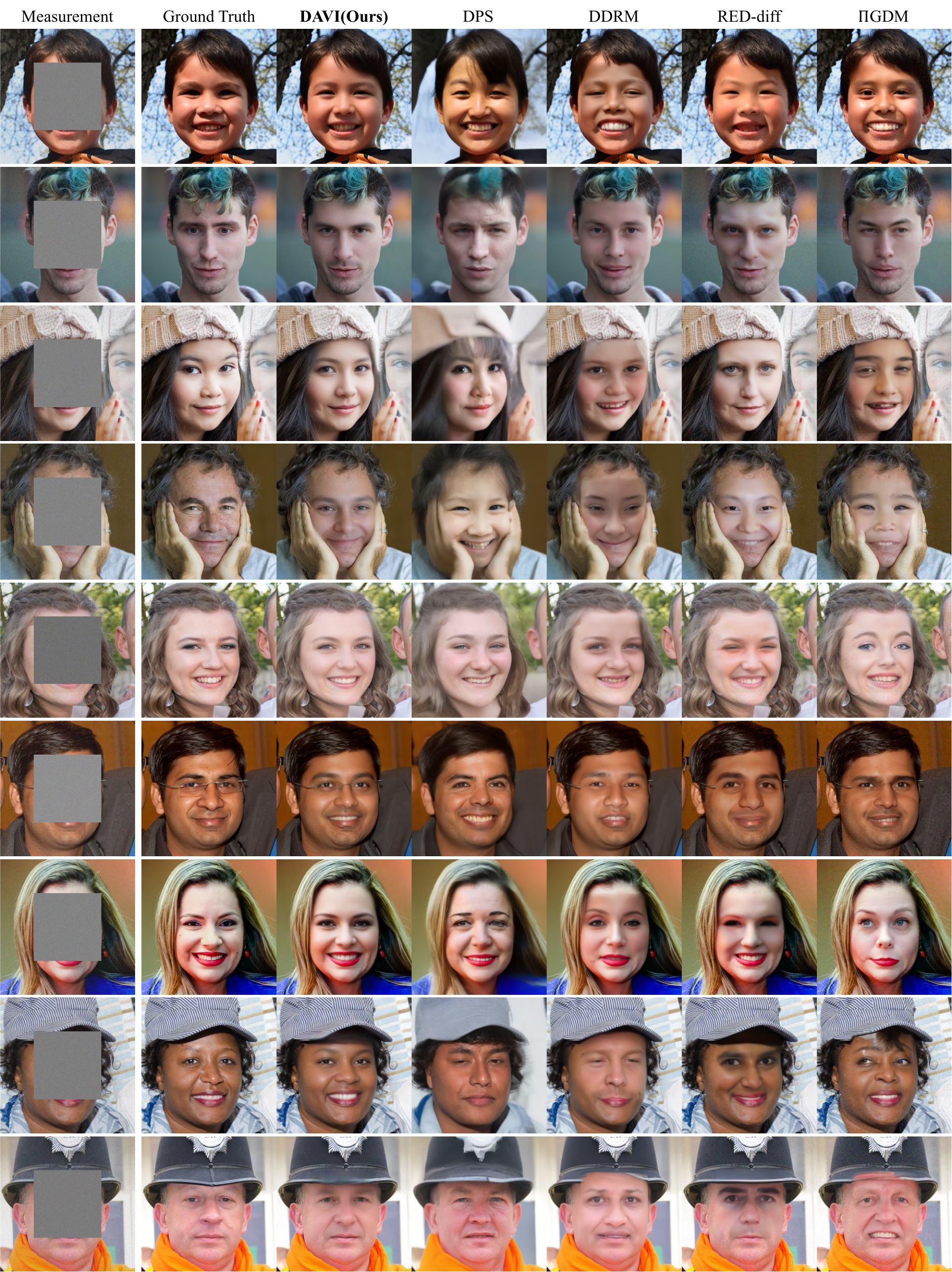}
    \caption{\textbf{Qualitative comparison of noisy inverse problem methods.} Results for box inpainting on FFHQ 256$\times$256.}
    \label{fig_supple:quality_ffhq_boxinpaint}
\end{figure*}

%% file: Figures_supple/quality_ffhq_deno.tex
\begin{figure*}[p]
    \centering
    \includegraphics[width=0.99\textwidth]{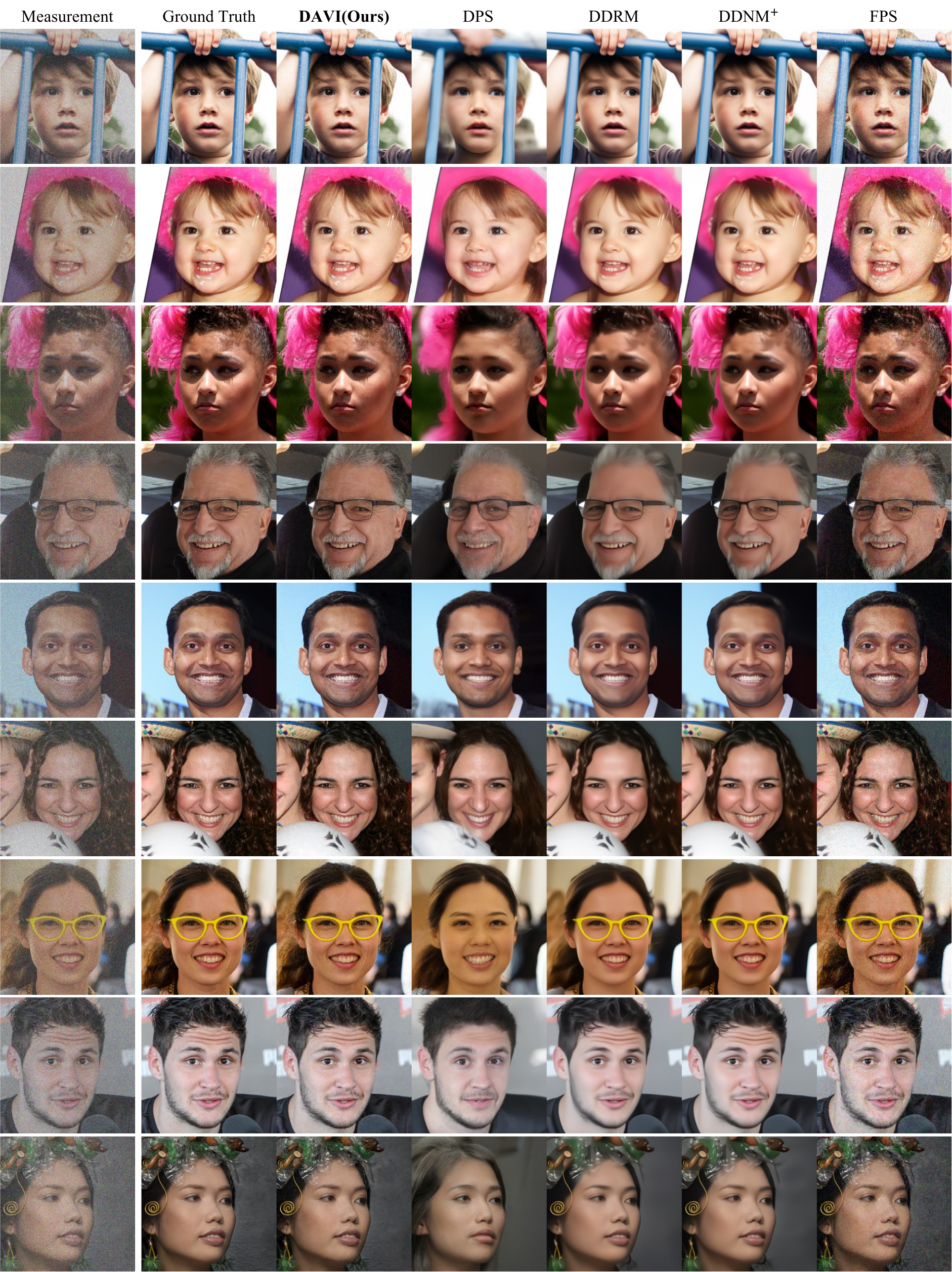}
    \caption{\textbf{Qualitative comparison of noisy inverse problem methods.} Results for denoising on FFHQ 256$\times$256.}
    \label{fig_supple:quality_ffhq_deno}
\end{figure*}

%% file: Figures_supple/quality_ffhq_color.tex
\begin{figure*}[p]
    \centering
    \includegraphics[width=0.99\textwidth]{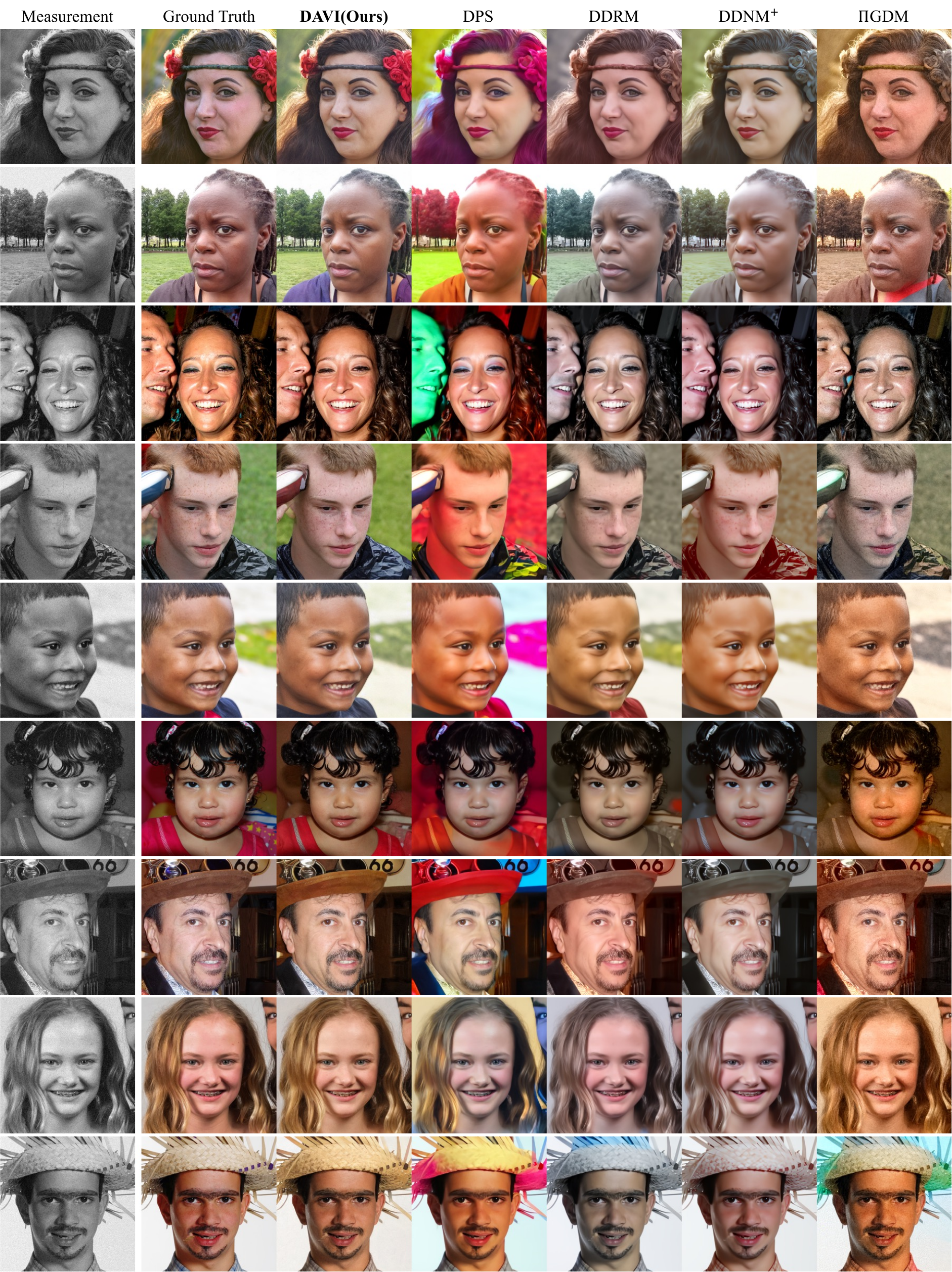}
    \caption{\textbf{Qualitative comparison of noisy inverse problem methods.} Results for colorization on FFHQ 256$\times$256.}
    \label{fig_supple:quality_ffhq_color}
\end{figure*}

%% file: Figures_supple/quality_imagenet_gauss.tex
\begin{figure*}[p]
    \centering
    \includegraphics[width=0.99\textwidth]{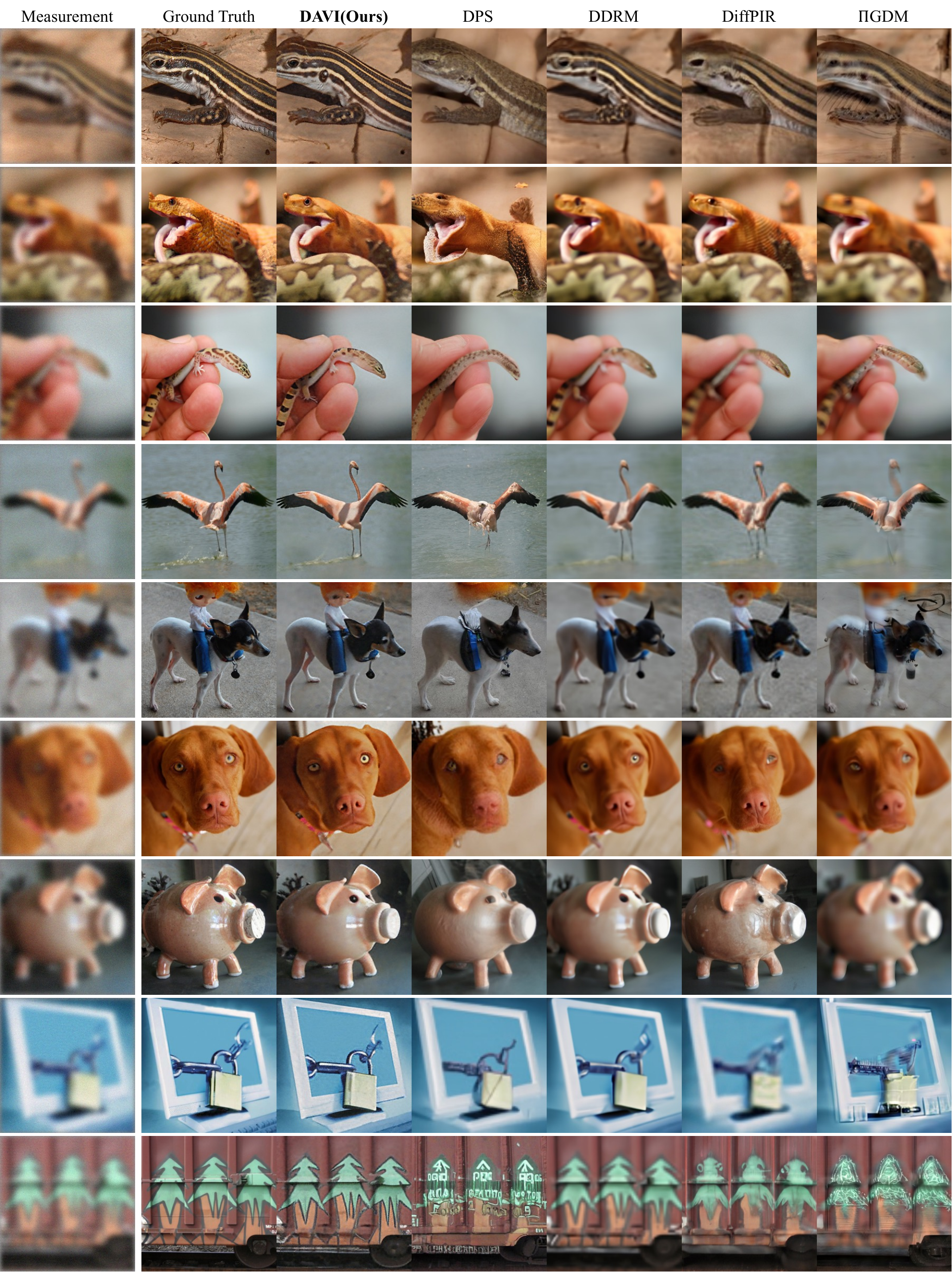}
    \caption{\textbf{Qualitative comparison of noisy inverse problem methods.} Results for Gaussian deblurring on ImageNet 256$\times$256.}
    \label{fig_supple:quality_imagenet_gauss}
\end{figure*}

%% file: Figures_supple/quality_imagenet_sr.tex
\begin{figure*}[p]
    \centering
    \includegraphics[width=0.99\textwidth]{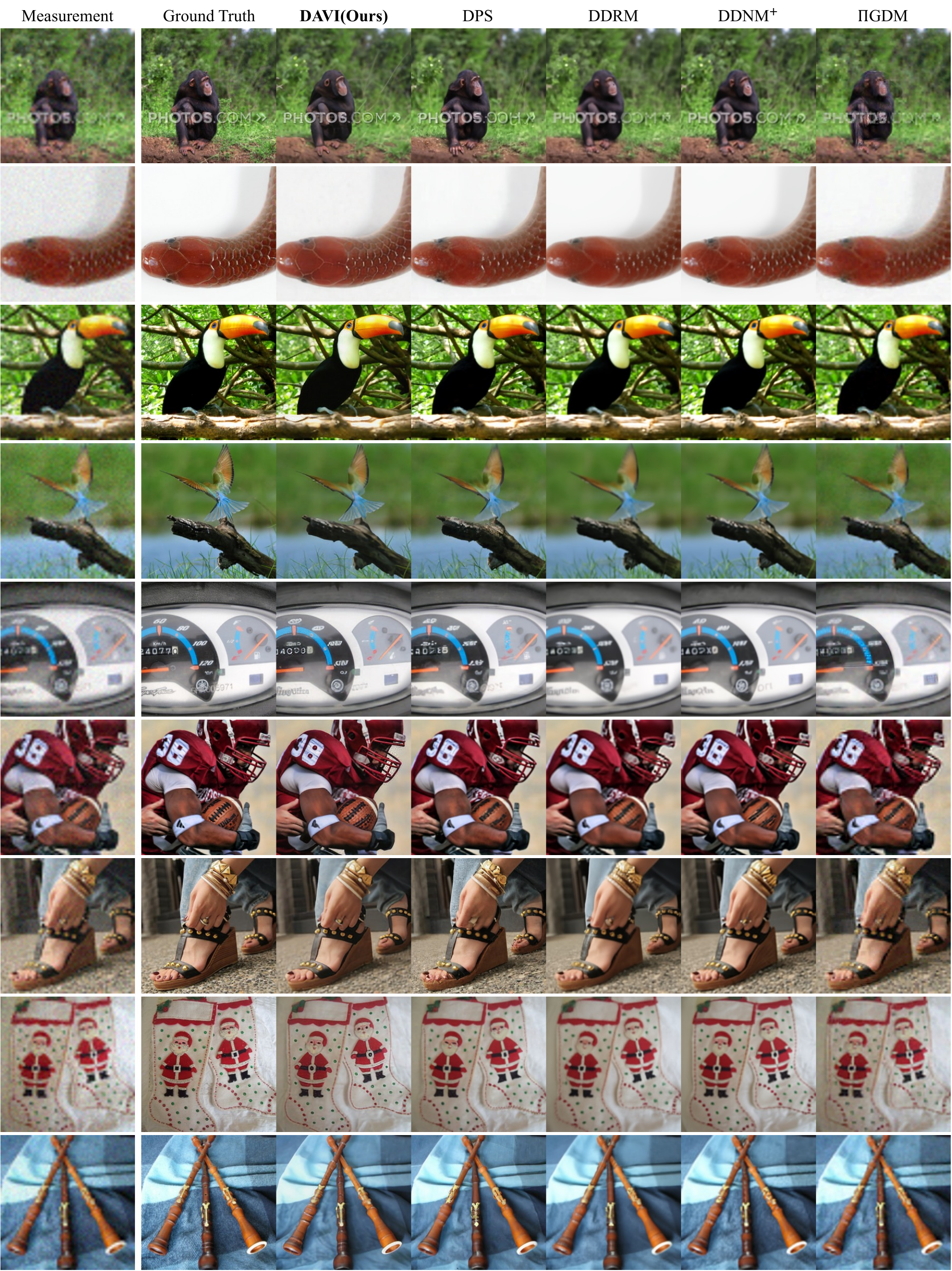}
    \caption{\textbf{Qualitative comparison of noisy inverse problem methods.} Results for 4$\times$ super-resolution on ImageNet 256$\times$256.}
    \label{fig_supple:quality_imagenet_sr}
\end{figure*}

%% file: Figures_supple/quality_imagenet_boxinpaint.tex
\begin{figure*}[p]
    \centering
    \includegraphics[width=0.99\textwidth]{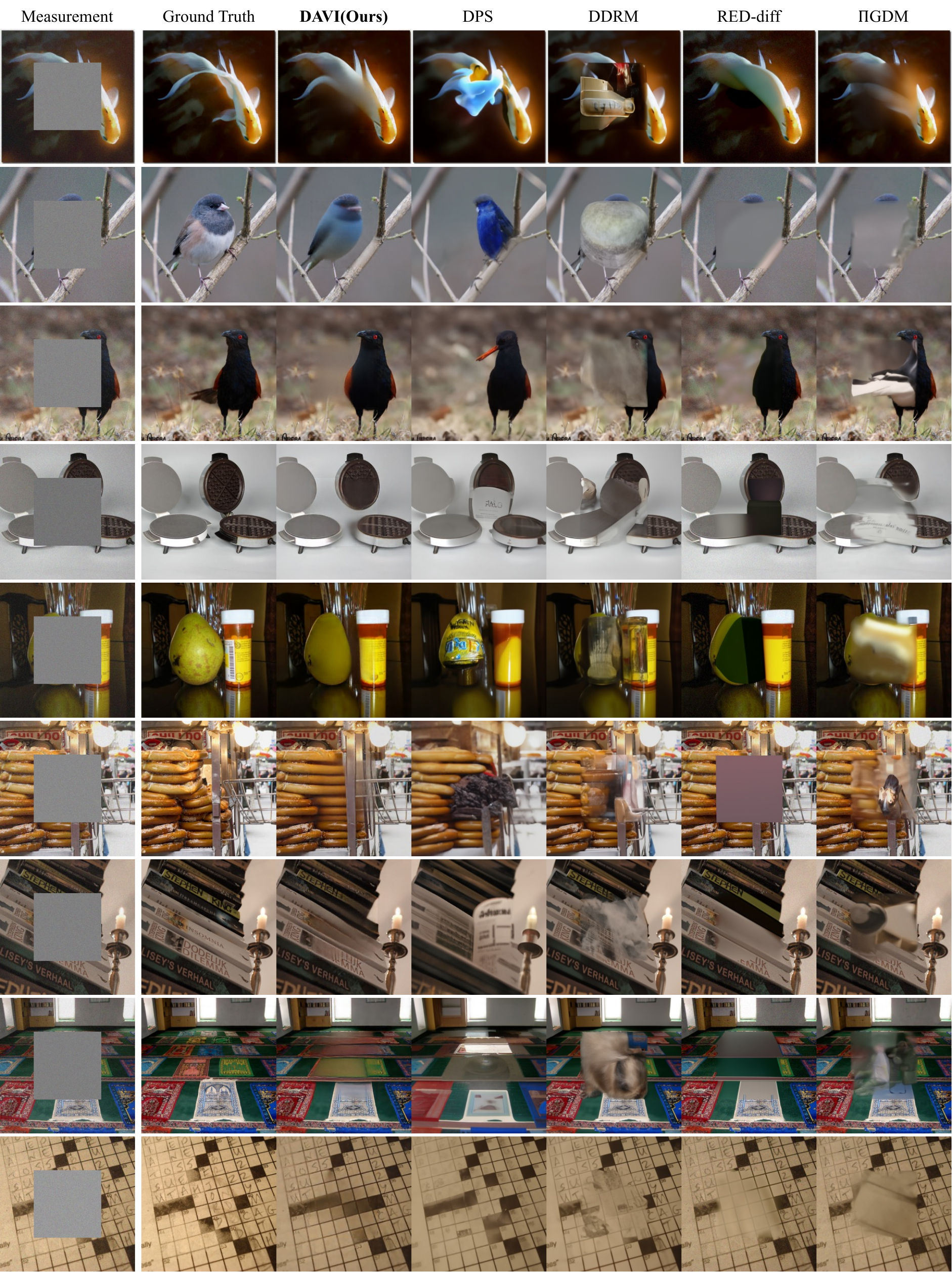}
    \caption{\textbf{Qualitative comparison of noisy inverse problem methods.} Results for box inpainting on ImageNet 256$\times$256.}
    \label{fig_supple:quality_imagenet_boxinpaint}
\end{figure*}

%% file: Figures_supple/quality_imagenet_deno.tex
\begin{figure*}[p]
    \centering
    \includegraphics[width=0.99\textwidth]{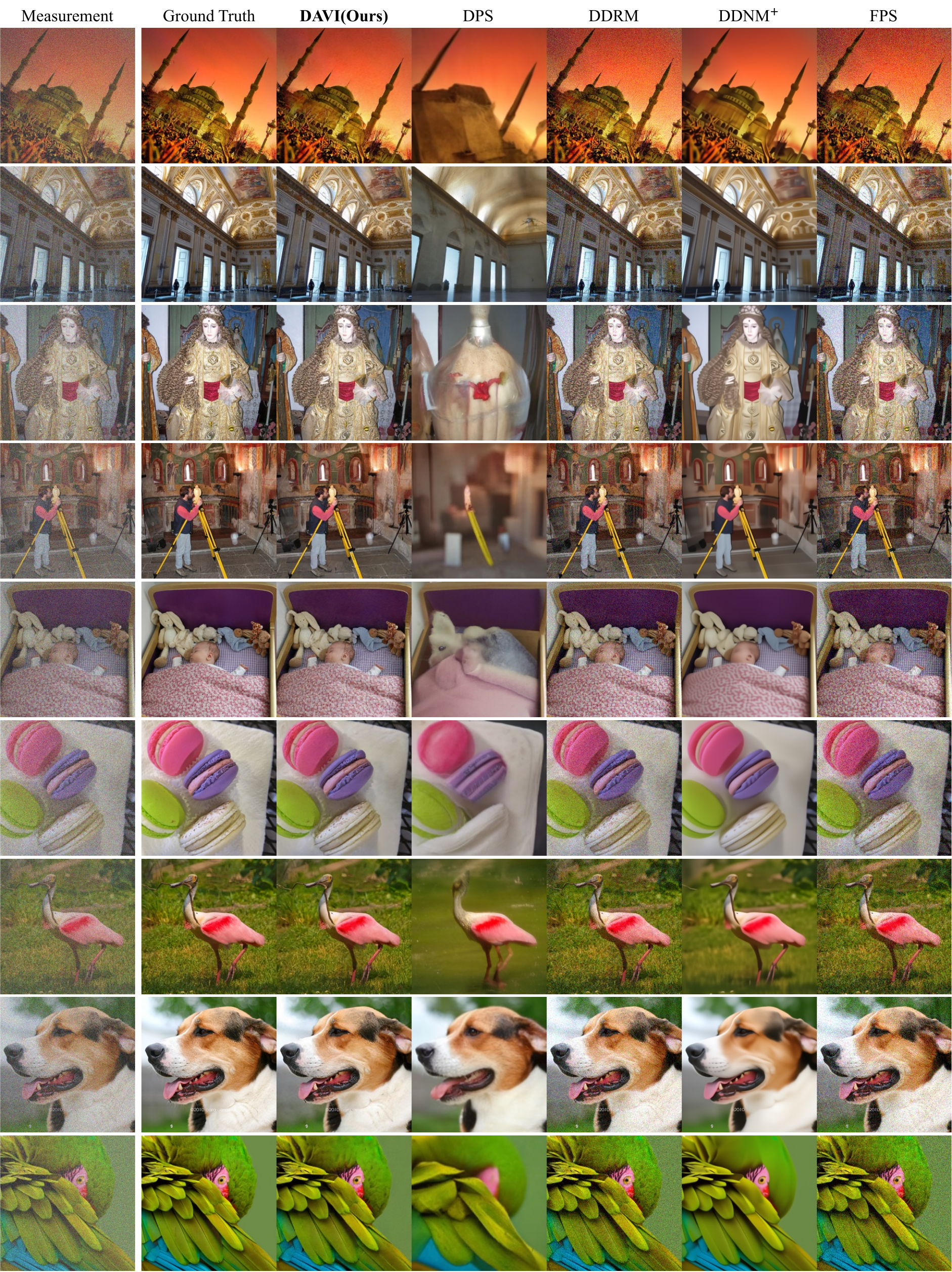}
    \caption{\textbf{Qualitative comparison of noisy inverse problem methods.} Results for denoising on ImageNet 256$\times$256.}
    \label{fig_supple:quality_imagenet_deno}
\end{figure*}

%% file: Figures_supple/quality_imagenet_color.tex
\begin{figure*}[p]
    \centering
    \includegraphics[width=0.99\textwidth]{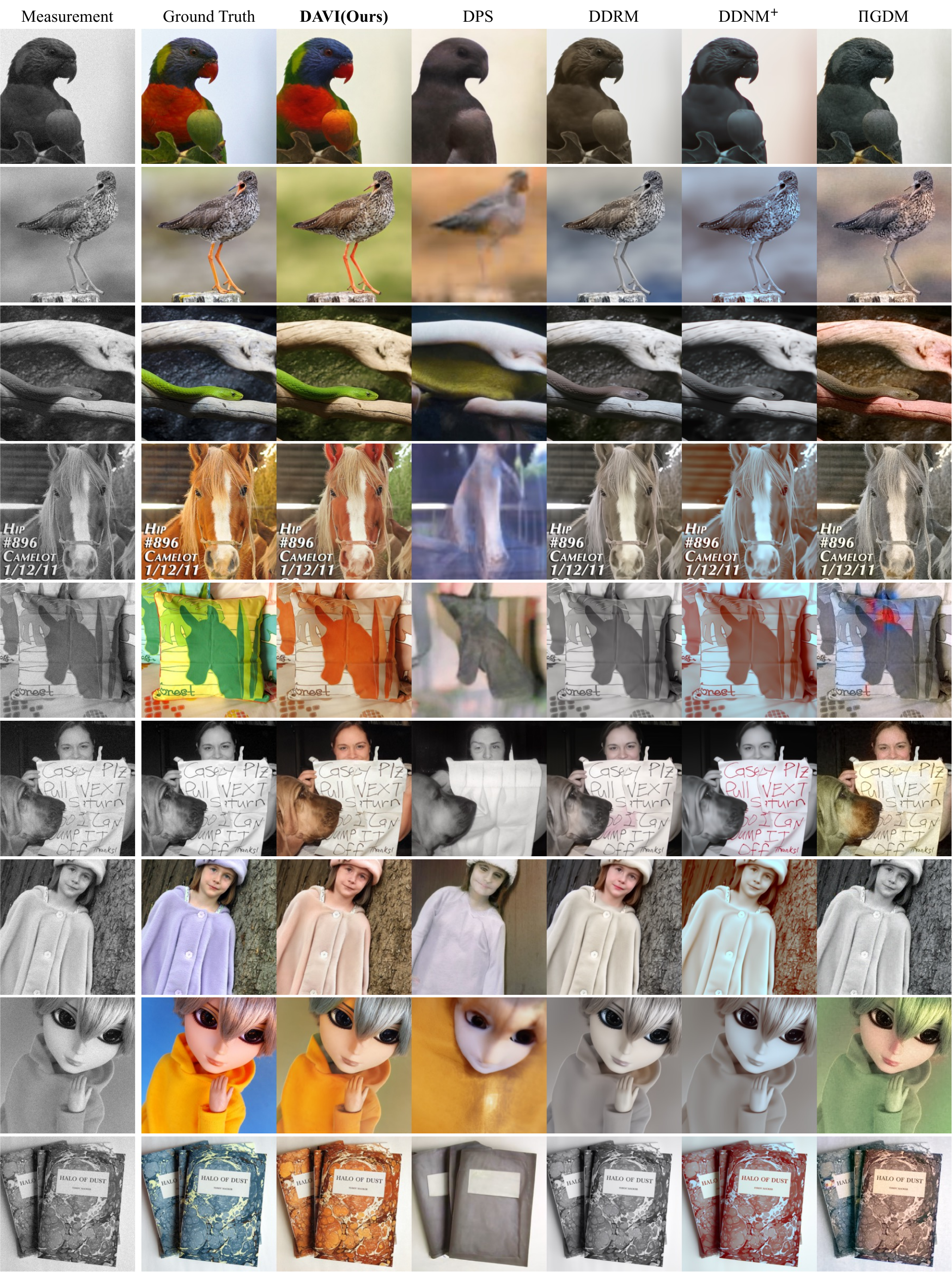}
    \caption{\textbf{Qualitative comparison of noisy inverse problem methods.} Results for colorization on ImageNet 256$\times$256.}
    \label{fig_supple:quality_imagenet_color}
\end{figure*}

%% file: main.bbl
\begin{thebibliography}{10}
\providecommand{\url}[1]{\texttt{#1}}
\providecommand{\urlprefix}{URL }
\providecommand{\doi}[1]{https://doi.org/#1}

\bibitem{anderson1982reverse}
Anderson, B.D.: Reverse-time diffusion equation models. In: Stochastic Processes and their Applications (1982)

\bibitem{chen2024videocrafter2}
Chen, H., Zhang, Y., Cun, X., Xia, M., Wang, X., Weng, C., Shan, Y.: Videocrafter2: Overcoming data limitations for high-quality video diffusion models. In: Conference on Computer Vision and Pattern Recognition) (2024)

\bibitem{choi2021ilvr}
Choi, J., Kim, S., Jeong, Y., Gwon, Y., Yoon, S.: {ILVR:} conditioning method for denoising diffusion probabilistic models. In: International Conference on Computer Vision (2021)

\bibitem{chung2023diffusion}
Chung, H., Kim, J., Mccann, M.T., Klasky, M.L., Ye, J.C.: Diffusion posterior sampling for general noisy inverse problems. In: International Conference on Learning Representations (2023)

\bibitem{chung2024direct}
Chung, H., Kim, J., Ye, J.C.: Direct diffusion bridge using data consistency for inverse problems. In: Advances in Neural Information Processing Systems (2024)

\bibitem{chung2022improving}
Chung, H., Sim, B., Ryu, D., Ye, J.C.: Improving diffusion models for inverse problems using manifold constraints. In: Advances in Neural Information Processing Systems (2022)

\bibitem{Chung_2022_CVPR}
Chung, H., Sim, B., Ye, J.C.: Come-closer-diffuse-faster: Accelerating conditional diffusion models for inverse problems through stochastic contraction. In: Conference on Computer Vision and Pattern Recognition (2022)

\bibitem{dhariwal2021diffusion}
Dhariwal, P., Nichol, A.: Diffusion models beat gans on image synthesis. In: Advances in neural information processing systems (2021)

\bibitem{dou2024diffusion}
Dou, Z., Song, Y.: Diffusion posterior sampling for linear inverse problem solving: A filtering perspective. In: International Conference on Learning Representations (2024)

\bibitem{ganguly2023amortized}
Ganguly, A., Jain, S., Watchareeruetai, U.: Amortized variational inference: A systematic review. In: Journal of Artificial Intelligence Research (2023)

\bibitem{gershman2014amortized}
Gershman, S., Goodman, N.: Amortized inference in probabilistic reasoning. In: Proceedings of the annual meeting of the cognitive science society (2014)

\bibitem{guilloteau2020hyperspectral}
Guilloteau, C., Oberlin, T., Bern{\'e}, O., Dobigeon, N.: Hyperspectral and multispectral image fusion under spectrally varying spatial blurs--application to high dimensional infrared astronomical imaging. In: IEEE Transactions on Computational Imaging (2020)

\bibitem{heusel2017gans}
Heusel, M., Ramsauer, H., Unterthiner, T., Nessler, B., Hochreiter, S.: Gans trained by a two time-scale update rule converge to a local nash equilibrium. In: Advances in neural information processing systems (2017)

\bibitem{ho2020denoising}
Ho, J., Jain, A., Abbeel, P.: Denoising diffusion probabilistic models. In: Advances in Neural Information Processing Systems (2020)

\bibitem{jordan1999introduction}
Jordan, M.I., Ghahramani, Z., Jaakkola, T.S., Saul, L.K.: An introduction to variational methods for graphical models. In: Machine learning (1999)

\bibitem{karras2017progressive}
Karras, T., Aila, T., Laine, S., Lehtinen, J.: Progressive growing of gans for improved quality, stability, and variation. In: arXiv preprint arXiv:1710.10196 (2017)

\bibitem{karras2022elucidating}
Karras, T., Aittala, M., Aila, T., Laine, S.: Elucidating the design space of diffusion-based generative models. In: Advances in Neural Information Processing Systems (2022)

\bibitem{karras2021alias}
Karras, T., Aittala, M., Laine, S., H{\"a}rk{\"o}nen, E., Hellsten, J., Lehtinen, J., Aila, T.: Alias-free generative adversarial networks. In: Advances in Neural Information Processing Systems (2021)

\bibitem{karras2019style}
Karras, T., Laine, S., Aila, T.: A style-based generator architecture for generative adversarial networks. In: Conference on Computer Vision and Pattern Recognition (2019)

\bibitem{kawar2022denoising}
Kawar, B., Elad, M., Ermon, S., Song, J.: Denoising diffusion restoration models. In: Advances in Neural Information Processing Systems (2022)

\bibitem{kingma2013auto}
Kingma, D.P., Welling, M.: Auto-encoding variational bayes. In: International Conference on Learning Representations (2014)

\bibitem{kostochastic}
Ko, J., Kong, I., Park, D., Kim, H.J.: Stochastic conditional diffusion models for robust semantic image synthesis. In: International Conference on Machine Learning (2024)

\bibitem{liu20232}
Liu, G.H., Vahdat, A., Huang, D.A., Theodorou, E.A., Nie, W., Anandkumar, A.: I$^{2}$sb: Image-to-image schrodinger bridge. In: International Conference on Machine Learning (2023)

\bibitem{luo2023diffinstruct}
Luo, W., Hu, T., Zhang, S., Sun, J., Li, Z., Zhang, Z.: Diff-instruct: A universal approach for transferring knowledge from pre-trained diffusion models. In: Advances in Neural Information Processing Systems (2023)

\bibitem{mardani2023variational}
Mardani, M., Song, J., Kautz, J., Vahdat, A.: A variational perspective on solving inverse problems with diffusion models. In: International Conference on Learning Representations (2023)

\bibitem{Mokady_2023_CVPR}
Mokady, R., Hertz, A., Aberman, K., Pritch, Y., Cohen-Or, D.: Null-text inversion for editing real images using guided diffusion models. In: Conference on Computer Vision and Pattern Recognition) (2023)

\bibitem{nah2017deep}
Nah, S., Hyun~Kim, T., Mu~Lee, K.: Deep multi-scale convolutional neural network for dynamic scene deblurring. In: Conference on Computer Vision and Pattern Recognition (2017)

\bibitem{ozbey2023unsupervised}
{\"O}zbey, M., Dalmaz, O., Dar, S.U., Bedel, H.A., {\"O}zturk, {\c{S}}., G{\"u}ng{\"o}r, A., {\c{C}}ukur, T.: Unsupervised medical image translation with adversarial diffusion models. In: IEEE Transactions on Medical Imaging (2023)

\bibitem{park2024ddmi}
Park, D., Kim, S., Lee, S., Kim, H.J.: Ddmi: Domain-agnostic latent diffusion models for synthesizing high-quality implicit neural representations. In: International Conference on Learning Representations (2024)

\bibitem{poole2022dreamfusion}
Poole, B., Jain, A., Barron, J.T., Mildenhall, B.: Dreamfusion: Text-to-3d using 2d diffusion. In: International Conference on Learning Representations (2023)

\bibitem{raj2023dreambooth3d}
Raj, A., Kaza, S., Poole, B., Niemeyer, M., Ruiz, N., Mildenhall, B., Zada, S., Aberman, K., Rubinstein, M., Barron, J., et~al.: Dreambooth3d: Subject-driven text-to-3d generation. In: International Conference on Computer Vision (2023)

\bibitem{richardson1972bayesian}
Richardson, W.H.: Bayesian-based iterative method of image restoration. In: JoSA (1972)

\bibitem{ILSVRC15}
Russakovsky, O., Deng, J., Su, H., Krause, J., Satheesh, S., Ma, S., Huang, Z., Karpathy, A., Khosla, A., Bernstein, M., Berg, A.C., Fei-Fei, L.: {ImageNet Large Scale Visual Recognition Challenge}. In: International Journal of Computer Vision (2015)

\bibitem{shu2018amortized}
Shu, R., Bui, H.H., Zhao, S., Kochenderfer, M.J., Ermon, S.: Amortized inference regularization. In: Advances in Neural Information Processing Systems (2018)

\bibitem{sohl2015deep}
Sohl-Dickstein, J., Weiss, E., Maheswaranathan, N., Ganguli, S.: Deep unsupervised learning using nonequilibrium thermodynamics. In: International Conference on Machine Learning (2015)

\bibitem{song2023pseudoinverse}
Song, J., Vahdat, A., Mardani, M., Kautz, J.: Pseudoinverse-guided diffusion models for inverse problems. In: International Conference on Learning Representations (2023)

\bibitem{song2019generative}
Song, Y., Ermon, S.: Generative modeling by estimating gradients of the data distribution. In: Advances in Neural Information Processing Systems (2019)

\bibitem{song2021solving}
Song, Y., Shen, L., Xing, L., Ermon, S.: Solving inverse problems in medical imaging with score-based generative models. In: International Conference on Learning Representations (2022)

\bibitem{song2020score}
Song, Y., Sohl{-}Dickstein, J., Kingma, D.P., Kumar, A., Ermon, S., Poole, B.: Score-based generative modeling through stochastic differential equations. In: International Conference on Learning Representations (2021)

\bibitem{steyvers2007probabilistic}
Steyvers, M., Griffiths, T.: Probabilistic topic models. Handbook of latent semantic analysis  (2007)

\bibitem{takeishi2021physics}
Takeishi, N., Kalousis, A.: Physics-integrated variational autoencoders for robust and interpretable generative modeling. In: Ranzato, M., Beygelzimer, A., Dauphin, Y.N., Liang, P., Vaughan, J.W. (eds.) Advances in Neural Information Processing Systems (2021)

\bibitem{vaswani2017attention}
Vaswani, A., Shazeer, N., Parmar, N., Uszkoreit, J., Jones, L., Gomez, A.N., Kaiser, {\L}., Polosukhin, I.: Attention is all you need. In: Advances in neural information processing systems (2017)

\bibitem{wang2022zero}
Wang, Y., Yu, J., Zhang, J.: Zero-shot image restoration using denoising diffusion null-space model. In: International Conference on Learning Representations (2023)

\bibitem{wang2023prolificdreamer}
Wang, Z., Lu, C., Wang, Y., Bao, F., Li, C., Su, H., Zhu, J.: Prolificdreamer: High-fidelity and diverse text-to-3d generation with variational score distillation. In: Advances in Neural Information Processing Systems (2023)

\bibitem{wei2024dreamvideo}
Wei, Y., Zhang, S., Qing, Z., Yuan, H., Liu, Z., Liu, Y., Zhang, Y., Zhou, J., Shan, H.: Dreamvideo: Composing your dream videos with customized subject and motion. In: Conference on Computer Vision and Pattern Recognition) (2024)

\bibitem{yin2024one}
Yin, T., Gharbi, M., Zhang, R., Shechtman, E., Durand, F., Freeman, W.T., Park, T.: One-step diffusion with distribution matching distillation. In: Conference on Computer Vision and Pattern Recognition (2024)

\bibitem{zamir2022restormer}
Zamir, S.W., Arora, A., Khan, S., Hayat, M., Khan, F.S., Yang, M.H.: Restormer: Efficient transformer for high-resolution image restoration. In: Conference on Computer Vision and Pattern Recognition (2022)

\bibitem{zhang2018unreasonable}
Zhang, R., Isola, P., Efros, A.A., Shechtman, E., Wang, O.: The unreasonable effectiveness of deep features as a perceptual metric. In: Conference on Computer Vision and Pattern Recognition (2018)

\bibitem{zhu2023denoising}
Zhu, Y., Zhang, K., Liang, J., Cao, J., Wen, B., Timofte, R., Gool, L.V.: Denoising diffusion models for plug-and-play image restoration. In: Conference on Computer Vision and Pattern Recognition Workshops (2023)

\end{thebibliography}
